\documentclass[10pt,twocolumn,letterpaper]{article}

\usepackage{cvpr}
\usepackage{times}
\usepackage{epsfig}
\usepackage{graphicx}
\usepackage{amsmath}
\usepackage{amssymb}

\graphicspath{{./figures/}}
\usepackage{bbold} % for a bold 1 (indicator function)
\usepackage{subfigure}
\usepackage{pgfplots}
\pgfplotsset{compat=1.3}  % ArXiv has a very outdated version
\usepackage{tikz}
\usepackage{pifont}
\usepackage{booktabs}
\usepackage{rotating}

\usepackage{siunitx}

\newcommand{\citep}[1]{\cite{#1}} %

\newcommand{\MPISintel}{MPI-Sintel}
\newcommand{\WHDRhinge}{WHDR-Hinge}

\def\margin{\xi}

\usepackage[breaklinks=true,bookmarks=false]{hyperref}

\usepackage[capitalise]{cleveref}  % nameinlink does not work with ArXiv

\cvprfinalcopy % *** Uncomment this line for the final submission

\def\cvprPaperID{3113} % *** Enter the CVPR Paper ID here

\begin{document}

%%%%%%%%% TITLE
\title{Reflectance Adaptive Filtering Improves Intrinsic Image Estimation}

%% Authors at the same institution
%%\author{First Author \hspace{2cm} Second Author \\
%%Institution1\\
%%{\tt\small firstauthor@i1.org}
%%}
%% Authors at different institutions
%\author{Thomas Nestmeyer\\
%Max Planck Institute for Intelligent Systems,\\
%T\"ubingen, Germany\\
%{\tt\small tnestmeyer@tuebingen.mpg.de
%% omit the following lines up until the closing ``}''.
%% Additional authors and addresses can be added with ``\and'',
%% just like the second author.
%% To save space, use either the email address or home page, not both
%\and
%Peter V. Gehler\\
%Bernstein Center for Computational Neuroscience,\\
%T\"ubingen, Germany, and \\
%MPI for Intelligent Systems, T\"ubingen, Germany\\
%{\tt\small pgehler@tuebingen.mpg.de}
%}

% for University of Wuerzburg affiliation:

% Authors at the same institution
%\author{First Author \hspace{2cm} Second Author \\
\author{Thomas Nestmeyer\textsuperscript{1,3} \hspace{2cm} Peter V. Gehler\textsuperscript{1,2,3}\\
\textsuperscript{1}University of W\"urzburg, Germany\\
\textsuperscript{2}Bernstein Center for Computational Neuroscience, T\"ubingen, Germany\\
\textsuperscript{3}Max Planck Institute for Intelligent Systems, T\"ubingen, Germany\\
{\tt\small \{tnestmeyer, pgehler\}@tuebingen.mpg.de
}
}

\maketitle
\thispagestyle{empty}

%% \renewcommand*{\thefootnote}{\fnsymbol{footnote}}
%% %\footnotetext{\textsuperscript{*} This work was performed while the authors were with the Max Planck Institute for Intelligent Systems, T\"ubingen, Germany. P.~V.~Gehler was additionally with the Bernstein Center for Computational Neuroscience, T\"ubingen, Germany.}
%% \footnotetext{\textsuperscript{*} This work was performed while P.~V.~Gehler was additionally with the Bernstein Center for Computational Neuroscience, T\"ubingen, Germany.}
%% \renewcommand*{\thefootnote}{\arabic{footnote}}
%% \setcounter{footnote}{0}

%%%%%%%%% ABSTRACT
\begin{abstract}
Separating an image into reflectance and shading layers poses a challenge for learning approaches because no large corpus of precise and realistic ground truth decompositions exists.
The Intrinsic Images in the Wild~(IIW) dataset provides a sparse set of relative human reflectance judgments, which serves as a standard benchmark for intrinsic images.
A number of methods use IIW to learn statistical dependencies between the images and their reflectance layer.
Although learning plays an important role for high performance, we show that a standard signal processing technique achieves performance on par with current state-of-the-art.
We propose a loss function for CNN learning of dense reflectance predictions.
Our results show a simple pixel-wise decision, without any context or prior knowledge, is sufficient to provide a strong baseline on IIW.
This sets a competitive baseline which only two other approaches surpass.
We then develop a joint bilateral filtering method that implements strong prior knowledge about reflectance constancy.
This filtering operation can be applied to any intrinsic image algorithm and we improve several previous results achieving a new state-of-the-art on IIW.
Our findings suggest that the effect of learning-based approaches may have been over-estimated so far. Explicit prior knowledge is still at least as important to obtain high performance in intrinsic image decompositions.
%\keywords{Intrinsic Images, Intrinsic Videos, Weak Labels, Reflectance and Shading}
\end{abstract}

%%%%%%%%% BODY TEXT
%%%%%%%%%%%%%%%%%%%%%%%%%%%%%%%%%%%%%%%%%%%%%%%%%%%%%%%%%%%%%%%%%%
\section{Introduction}\label{sec:introduction}
% Set the stage, intrinsic images are important
Almost 40 years ago, the seminal paper of Barrow and Tenenbaum~\cite{barrow1978intrinsic} conjectured that \emph{``A robust visual system should be organized around a noncognitive, nonpurposive level of processing that attempts to recover an intrinsic description of the scene''}.
Their work motivates the task of decomposing an image into constituent layers such as surface reflectance, surface orientation, distance and incident illumination.
Ever since, significant progress has been made on this problem, but the recovery of these physical properties of visual scenes or videos remains an open challenge.
A successful model needs to resolve the ill-posedness of the problem and cope with the variety of image appearances.

% Now CNNs enter the scene
A possible line of attack are supervised learning methods which have been used with great success for a wide range of computer vision applications.
Standing out for superior performance combined with favorable runtimes is the class of Convolutional Neural Network (CNNs), a dominant contender for many vision problems.
CNNs are mostly falling into the category of purposive models, guided by task specific goals such as image classification or recognition.
The obvious question is whether CNNs will fare equally well on the problem of intrinsic image decompositions.

% what has been done.
Several works have included CNN methods in systems that recover reflectance and shading layers~\cite{narihira2015directIntrinsics,narihira2015lightness,zhou2015learningPriors,zoran2015ordinal}.
However, prior work uses CNNs mostly in combination with additional methods, such as Conditional Random Fields (CRFs), to achieve a dense image decomposition.
An advantage of CRF models is their ability to encode prior information about the problem.
In the \mbox{pre-CNN} time, intrinsic image methods were dominated by CRF models with carefully designed priors on reflectance, shading, and their combination.
In this paper we attempt to answer the question whether prior terms are necessary when human annotation in the form of weak labels is available.

% Where does training data come from?
Acquiring accurate training data for intrinsic images is a challenge.
The MIT intrinsic dataset~\cite{grosse2009mitIntrinsic} with 20 images and 10 (single color) light configurations was a first attempt to empirically validate intrinsic estimation techniques.
It has served this purpose well, but lacks realism and diversity.
Recently,~\cite{beigpour2015multiIlluminantDataset} proposed an extension to multi illuminants, but without overcoming the limitations on extent.
Another possible route to generate datasets is the use of computer graphics rendering engines.
This has been explored by the authors of~\cite{beigpour2013synthetic} who created a dataset of synthetic scenes rendered using the Blender open source rendering engine~\cite{blender}.
This led to a dataset of 32 single objects and 36 scene compositions which is still limited in terms of detail and diversity.
The \MPISintel~dataset has been created using the open source movie Sintel~\cite{butler2012sintel} to serve as a benchmark for several problems such as optical flow estimation.
While \MPISintel~is more varied and complex, the type of scenes and visual appearance is still very different from real world data.

% Now Bell comes along
A significant attempt to overcome the lack of empirical data is the dataset of ``Intrinsic Images in the Wild''~(IIW)~\cite{bell2014iiw}.
This dataset contains 5230 photos of mostly indoor scenes which have been annotated with a sparse set of relative reflectance judgments.
From a small set of image locations, human judgments on pairs of neighboring locations have been collected, which provide whether one point is of darker or similar material reflectance.
Although humans can be fooled with artificial setups~\citep{adelson2000illusions}, the perception of relative material reflectance is sufficient to provide mostly consistent label information for this large corpus of images (see~\cite{bell2014iiw} for an analysis).
%. \tn{consistent: \cite{bell2014iiw} says 92.8\% triangle consistency, but the wrong ones are the ones with low confidence}
% \tn{Should we also mention something more specific to human perception? Like the paper from Kingdom: Lightness, brightness and transparency: A quarter century of new ideas, captivating demonstrations and unrelenting controversy}
Along with the dataset, Bell et al.~also formulate a performance metric (WHDR) that we will discuss in detail in~\cref{sec:whdr}.
The IIW dataset allows to empirically validate intrinsic image estimation and the judgments have also been used to train models for intrinsic image decompositions~\cite{narihira2015lightness,zhou2015learningPriors,zoran2015ordinal}.

% What are we up to?
In this paper we develop two intrinsic image models: a CNN approach with appropriate loss function and a filtering technique to include strong prior knowledge about reflectance properties.
%In this paper we develop a model that directly infers dense reflectance and shading using weak label information from the IIW dataset as training information.
We first design a CNN method that, in contrast to previous work, does not include prior information on shading smoothness~\cite{land1971retinex}, reflectance~\cite{omer2004colorlines,gehler2011sparsity,shen2011sparse,bi2015l1}, or combinations~\cite{barron2015sirfs}.
We design a loss function that enables end-to-end learning from the pairwise judgments.
This leads to an interesting result: a simple multi-layer perceptron with no image context, just based on the pixels alone provides competitive performance, better or on par with current learning and non-learning models.
We then develop a method from the other extreme, a dense filtering operation based on joint bilateral and guided filtering.
This technique simplifies the processing pipeline of~\cite{bi2015l1} and makes it possible to apply to any reflectance prediction.
Our experiments show drastically improved state-of-the-art performance on~IIW.
Besides presenting the empirically best performing algorithm, our results reveal interesting observations about the current state of intrinsic image estimation.
%% A) A simple context free, per pixel CNN decision performs better than most prior methods, this sets a new most previous algorithms, setting a new bar.
%% B) The filtering step improves performance drastically, it is this inclusion of prior knowledge that we found to drastically improve performance.
In summary, we believe that for intrinsic image estimation, it is the inclusion of prior knowledge through regularization, CRFs, or filtering that still drives the performance.
To rely solely on learning approaches, the amount of available annotation may still be insufficient.

%%%%%%%%%%%%%%%%%%%%%%%%%%%%%%%%%%%%%%%%%%%%%%%%%%%%%%%%%%%%%%%%%%
\section{Related Work}
% Intrinsic images are useful because of: shadow removal~ \cite{kumar2011shadowRemoval}, robotics, self driving cars~\cite{maddern2014autonomous}, Object Recoloring~\cite{beigpour2011objectRecoloring}, in general for image editing, Interactive Intrinsic image video editing~\cite{bonneel2014interactiveVideoEditing}

Until recently there was a lack of empirical data to validate intrinsic images algorithms. Therefore, most of the literature
revolved around the design of suitable priors. The recent work of~\cite{barron2015sirfs} is a prominent example of a
method that carefully trades the use of prior information with interesting representations that enable a detailed decomposition into several layers.
Priors in~\cite{barron2015sirfs} include terms on smoothness, parsimony and absolute values of reflectance, smoothness, surface isotropy and occluding contour priors on shape and a multivariate Gaussian fit to a spherical-harmonic illumination model.
This lead to impressive results on the MIT intrinsic dataset~\cite{grosse2009mitIntrinsic}, but the method is limited to single masked objects in a scene, and problems with complex illumination remain.

% alternative with filtering and very simple prior
The work of~\cite{bi2015l1} approaches the problem from a filtering perspective.
After a filtering step followed by clustering, the pixels are grouped into regions of same reflectance, such that a simple shading term suffices to recover the full intrinsic decomposition. This method produces the best results on the IIW dataset but takes several minutes of processing time. In~\cref{sec:filtering} we build on this work and propose a filtering technique that can be applied to any other intrinsic image estimation as well. This implements the idea of grouping pixels into sets of constant reflectance.
% not only single RGB as input
Other works consider additional knowledge in order to recover reflectance and shading, as, \eg, multiple images of the same scene with different lighting~\cite{weiss2001sequences,laffont2015sequences}, an interactive setting with user annotations~\cite{bousseau2009userAssisted,bonneel2014interactiveVideoEditing}, or an additional depth layer as input~\cite{chen2013depth}.

The paper of~\cite{bell2014iiw} introduced the Intrinsic Images in the Wild dataset with human annotations giving relative reflectance judgments that served as the training and test set for different learning based methods. Using this data, the work of~\cite{bell2014iiw} was the first to compare different algorithms on a large corpus of real world scenes.

% predicting only judgments
A first attempt to learn using the data from IIW was made by~\cite{narihira2015lightness}. The authors used the relative judgment information in a multi-class setup and fine-tune an AlexNet CNN trained on ImageNet. Only the sparse annotation points that are required to compute the WHDR loss are predicted with this network and there is no step that turns them into a dense decomposition.
% CNN plus CRF
The works of~\cite{zhou2015learningPriors} and~\cite{zoran2015ordinal} are similar, both use a CNN to obtain pairwise judgment predictions, then followed by a step to turn
the sparse information into a dense decomposition.
Both methods achieve good results on IIW and take several seconds to process an image.

% CNN only
Similar to our work, in the sense that a dense intrinsic decomposition is predicted, is the work of~\cite{narihira2015directIntrinsics}.
A CNN is used to directly predict reflectance and shading with the objective function being the difference to ground truth decompositions.
Since those are only available for the rendered dataset of~\MPISintel{}, the authors report that the learned model does not generalize well to the real world images of~IIW.
An additional data term in the gradient domain is used by~\cite{lettry2016darn}. They also propose to use an adversary in order to remove typical generative CNN artifacts by discriminating between generated and ground truth decompositions.
Therefore, this approach has the same limitation requiring dense ground truth decompositions and no results on IIW are available.
To our knowledge, there is no CNN based method that predicts a dense intrinsic decomposition and works well for images from IIW.

In~\cref{tab:methodComparison} we organized the related work along the dimensions that are relevant for the proposed method.

The work of~\cite{chen2016depthInTheWild} also trains a CNN from relative judgments with a ranking loss to predict pixel-wise labels, but for the application of recovering dense depth estimates. This involved the creation of a dataset with relative depth judgments in the spirit of IIW. However, in contrast to intrinsic images, it is possible to capture accurate ground truth depth for training and testing, making reflectance and shading estimation a more relevant target of learning from sparse pairwise comparisons.

% to choose a checkmark and cross out, look at the following possibilities:
% \ding{51}\ding{52}
% \ding{54}\ding{55}\ding{56}
% for now I chose those:
\def\yes{\ding{52}}
\def\no{\ding{56}}
\newcommand{\kw}[1]{\emph{#1}}  % write keywords in table italic

\newcolumntype{C}{>{\centering\let\newline\\\arraybackslash\hspace{0pt}}m{13mm}}
\newcolumntype{D}{>{\centering\let\newline\\\arraybackslash\hspace{0pt}}m{17mm}}
\begin{table}%[t]
\caption{Overview of different intrinsic image estimation methods. For every method we note whether or not it uses a \kw{CNN trained on IIW} and whether the \kw{CNN decomposes densely} into intrinsic layers without an additional globalization step.}
\begin{tabular}{lCD}  % change width of C in newcolumntype above
Method                & CNN trained on IIW  & CNN \mbox{decomposes} densely\\%direct dense decompos. by CNN\\
% method              & CNN  & full \\
\hline
Retinex~\cite{land1971retinex}  & \no  & \no  \\
% Shen et al. 2011 - CVPR: Intrinsic Images Decomposition Using a Local and Global Sparse Representation of Reflectance. Uses a reflectance sparsity prior
%Shen et al. 2011~\cite{shen2011sparse}                          & \no  & \yes & \no     & n/a  \\
% Garces et al. 2012 - CGF: Intrinsic Images by Clustering. Based on simple assumptions about its reflectance and luminance
%Garces et al. 2012~\cite{garces2012clustering}                  & \no  & \yes & \no     & n/a  \\
% Zhao et al. 2012 - PAMI: A Closed-form Solution to Retinex with Non-local Texture Constraints.
%Zhao et al. 2012~\cite{zhao2012nonlocalTextureConstraints}      & \no  & \yes & \no     & n/a  \\
% Bell et al. 2014 - SIGGRAPH: Intrinsic Images in the Wild
Bell et al. 2014~\cite{bell2014iiw}  & \no  & \no \\
% Bonneel et al. 2014 - SIGGRAPH: Interactive Intrinsic Video Editing
%Bonneel et al. 2014~\cite{bonneel2014interactiveVideoEditing}   & \no  & \yes & \no     & n/a  \\
% Bi et al. 2015 - SIGGRAPH: An L1 image transform for edge-preserving smoothing and scene-level intrinsic decomposition
Bi et al. 2015~\cite{bi2015l1}  & \no  & \no \\
% Narihira et al. 2015 - CVPR: Learning lightness from human judgement on relative reflectance
Narihira et al. 2015a~\cite{narihira2015lightness}      & \yes & \no \\
% Zhou et al. 2015 - ICCV: Learning data-driven reflectance priors for intrinsic image decomposition
Zhou et al. 2015~\cite{zhou2015learningPriors}              & \yes & \no \\
% Zoran et al. 2015 - CVPR: Learning ordinal relationships for mid-level vision: solve a constrained quadratic optimization problem and therefore is non-CRF? still needs a prior
Zoran et al. 2015~\cite{zoran2015ordinal}             & \yes & \no\\
% Narihira et al. 2015 - ICCV: Direct intrinsics: Learning albedo-shading decomposition by convolutional regression
Narihira et al. 2015b~\cite{narihira2015directIntrinsics}      & \no & \yes   \\
% Narihira et al. 2016 - ArXiv: DARN: a Deep Adversial Residual Network for Intrinsic Image Decomposition
Lettry et al. 2016~\cite{lettry2016darn}     & \no & \yes   \\
% ours:
\textbf{This paper}  & \yes & \yes
\end{tabular}
\label{tab:methodComparison}
\end{table}

%%%%%%%%%%%%%%%%%%%%%%%%%%%%%%%%%%%%%%%%%%%%%%%%%%%%%%%%%%%%%%%%%%
\section{Preliminaries}\label{sec:prelim}
We work with linear RGB and the Lambertian reflectance assumption, which allows to separate every pixel in image $I\in [0,1]^{3 \times h \times w}$ into a product of reflectance~$R$ and shading~$S$, that is the pixel and channel-wise product $I=RS$.
Further, we assume achromatic light which reduces the decomposition problem to a per-pixel scalar estimation problem.
Namely, given a scalar $r_p\in[0,1]$ for each pixel $p$, we recover reflectance and shading as
\begin{align}
% \forall i,c:\exists s_i:\
% I_p^c = r_p\frac{I_p^c}{\Vert I_p\Vert} \cdot \frac{1}{r_p}\Vert I_p\Vert  % estimate r with ||I||
% I_p^c = \frac{1}{s_p}\frac{I_p^c}{\Vert I_p\Vert} \cdot s_p\Vert I_p\Vert  % estimate s with ||I||
% R_p^c = r_p\cdot\frac{I_p^c}{\frac{1}{3}\sum_c{I_p^c}}, \qquad S_p^c = \frac{1}{r_p}\cdot\frac{1}{3}\sum_c{I_p^c}  % estimate r with R+G+B
% R_p = r_p\cdot\frac{1}{\frac{1}{3}\sum_c{I_p^c}}\cdot I_p, \qquad S_p = \frac{1}{r_p}\cdot\frac{1}{3}\sum_c{I_p^c}\cdot\begin{pmatrix}1,&1,&1\end{pmatrix}  % estimate r with R+G+B
R_p = \frac{r_p}{\frac{1}{3}\sum_{c}{I_p^c}}\cdot I_p, \qquad S_p = \frac{\frac{1}{3}\sum_c{I_p^c}}{r_p}\cdot\begin{pmatrix}1\\1\\1\end{pmatrix},  % estimate r with R+G+B
\label{eq:recover_reflectance_shading}
\end{align}
where $c\in\{R,G,B\}$ denotes the color channel.
Under these assumptions, the problem boils down to estimation of a single scalar per pixel $\mathbf{r}\in\mathbb{R}^{h \times w}$.

The same assumptions are commonplace in the literature and have been used, \eg, in~\cite{gehler2011sparsity}.
We note that achromatic light is often violated in the IIW dataset, especially in the presence of multiple light sources.
As the proposed loss function WHDR only compares relative lightness and no color information, it is invariant to this choice.

%%%%%%%%%%%%%%%%%%%%
\subsection{A quantitative measure for intrinsic images}\label{sec:whdr}
Accurate ground truth information in the form of image decompositions in reflectance and shading layers does not exist at scale.
To empirically validate the quality of intrinsic image algorithms using the pairs of relative reflectance judgments alone, Bell et al.~\cite{bell2014iiw} introduced the WHDR metric (weighted human disagreement rate).
We refer to their work for all details on the data annotation process, but will review the ingredients that we need for our development.

For every image, annotation is given in the form of pairs of image locations $(i_1,i_2)$ for which a human reflectance judgment $J_i\in\{1,2,E\}$ is provided.
The judgment indicates whether point $i_1$ is darker than $i_2$ ($J_i=1$), lighter ($J_i=2$), or of equal reflectance ($J_i=E$).
The confidence $w_i$ of a judgment is defined via the CUBAM score of the two-decision model ``points have the same reflectance'' and if not ``does the darker point have darker reflectance'' (see~\cite{bell2014iiw} for further details).
The annotation set $\{(i_1,i_2,J_i,w_i)\}_{i=1,\ldots,N_I}$ varies in size $N_I$ for every image~$I$ in the range from 1 to 1181 with a median of~113.

Given a reflectance prediction $R$, first a relative classification for the set of annotated points is computed as
%\begin{align}
%\hat J_{\delta}(R,i) &= \begin{cases}
%1 & \text{if } R_{q_i}/R_{p_i} > 1+\delta\\
%2 & \text{if } R_{p_i}/R_{q_i} > 1+\delta\\
%E & \text{else},
%\end{cases}\label{eq:judgmentRewritten}
%\end{align}
\begin{align}
\hat J_{\delta}(R,i) &= \begin{cases}
1 & \text{if } R_{i_2}/R_{i_1} > 1+\delta\\
2 & \text{if } R_{i_1}/R_{i_2} > 1+\delta\\
E & \text{else},
\end{cases}\label{eq:judgmentRewritten}
\end{align}
where $\delta\geq 0$ controls when two points are considered different.
%\tn{use L instead of R for reflectance lightness in this formula, since otherwise not clear what it should mean. \cite{bell2014iiw} uses lightness as the mean of RGB (and therefore ignores perceptual differences in the color channels when computing WHDR (defined later) for this classification).}
For large values of $\delta$, two points would need to be farther apart to be judged as darker (resp. lighter).

Given these relative estimates, the WHDR loss is computed as the weighted average of how often the annotation and prediction disagree
\begin{align}
% \text{WHDR}_\delta(J, R) &= \frac{\sum_{i} w_i\cdot \mathbb{1}\left(J_i\neq\hat J_{i,\delta}(R)\right)}{\sum_{i} w_i}
\text{WHDR}_\delta(J, R) &= \frac{\sum_{i} w_i\cdot \mathbb{1}\left(J_i\neq\hat J_{\delta}(R,i)\right)}{\sum_{i} w_i}.
\label{eq:whdr}
\end{align}
Note that this loss does not evaluate the reflectance at all points in the image, but only at those for which labels are available.
Therefore, it could also be evaluated on these points alone for an algorithm that does not provide a dense decomposition of the image.

The works of~\cite{narihira2015lightness,zhou2015learningPriors,zoran2015ordinal} use these relative annotations to train multi-class classifiers, predicting for every pair of patches its relative reflectance judgment~$\{1,2,E\}$ directly.
Since this approach does not provide the actual values $R$ of the reflectance layer, further post-processing steps are required to produce a dense prediction.
These post-processing steps are separate from the classifiers and motivated by common intrinsic prior terms.
We will circumvent any post-processing by directly predicting a dense reflectance map~$R$.

%%%%%%%%%%%%%%%%%%%%%%%%%%%%%%%%%%%%%%%%%%%%%%%%%%%%%%%%%%%%%%%%%%
\section{Direct Reflectance Prediction with a CNN}\label{sec:direct_CNN_prediction}
We propose an objective function that makes direct use of the relative reflectance judgments by humans that the IIW dataset provides.
This weak label information has been used in~\cite{zhou2015learningPriors,zoran2015ordinal} for CNN training already, treating it however as a multi-class classification problem.
While a multi-class loss achieves good performance on pairs of points, this strategy requires an additional globalization step to propagate information to all pixels.
Our aim is to directly decompose the entire image with a single forward pass of a CNN, avoiding any need for post-processing.

We will first discuss the loss function that we use and then describe the network architecture and training method.
%
%%%%%%%%%%%%%%%%%%%%
\subsection{\WHDRhinge{} loss}\label{sec:whdr_hinge}

% line width in the following plots:
\def\thickness{1.5}
\def\xmin{0}
\def\xmax{1.7}
\def\ymin{-0.2}
\def\ymax{1.5}
\def\paramDelta{0.1}
\def\paramMargin{0.075}
\def\height{37mm}
\def\width{0.9\linewidth}
\colorlet{darkgreen}{green!70!black}

\begingroup % to have change of next line only here in this group
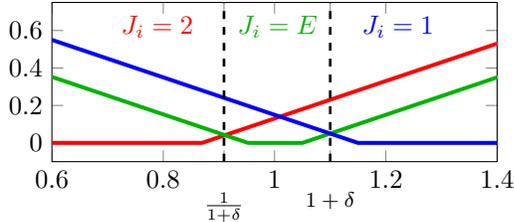
\begin{figure}[t]
%\includegraphics{tikz_figure_WHDRhinge}
%% \subfigure[]{\label{fig:WhdrHingeClass2}
%% \begin{tikzpicture}
%%   \pgfmathsetmacro{\deltaxRight}{(1+\paramDelta)}
%%   \pgfmathsetmacro{\lossZero}{(1+\paramDelta+\paramMargin)}
%%   \begin{axis}[
%%   	height=\height,
%%   	width=\width,
%%     %	samples=200,
%%   	xmin=0.25,
%%   	xmax=\xmax,
%%   	ymin=\ymin,
%%   	ymax=1.25,
%%     extra x ticks={\lossZero},
%%     extra x tick labels={$1+\delta+\margin$},
%%     extra x tick style={xticklabel style={yshift=-8, font=\footnotesize}}
%%     %xlabel=$\frac{R_{1,i}}{R_{2,i}}$,
%%     % ylabel={$\max(0, \margin - (x - \delta_x))$}
%%   ]
%% 	  % hinge loss
%%     % \addplot+[blue, mark=none]{max(0, \paramMargin - (x - \deltaxRight))};
%%     % hinge loss as connection of line segments (check that it is the same)
%%     \addplot [blue, line width=\thickness] coordinates { (1+\paramDelta+\paramMargin-2, 2) (1+\paramDelta+\paramMargin, 0) (4, 0) };

%%     % \addplot [dashed] coordinates { (-3,\paramMargin) (\deltaxRight,\paramMargin) };
%%     % \addplot [dashed] coordinates { (\deltaxRight,-1) (\deltaxRight,\paramMargin) };
%%     \node[label={90:$1+\delta$}] at (axis cs:\deltaxRight,1) {};
%%     % \node[label={180:$\margin$}] at (axis cs:0,\paramMargin) {};
%%     % correct loss
%%     \addplot [dashed, line width=\thickness] coordinates { (-3,1) (\deltaxRight,1) };
%%     \addplot [dashed, line width=\thickness] coordinates { (\deltaxRight,1) (\deltaxRight,0) };
%%     \addplot [dashed, line width=\thickness] coordinates { (\deltaxRight,0) (5,0) };
%%   \end{axis}
%% \end{tikzpicture}
  %% }
\def\xmin{0.6}
\def\xmax{1.4}
\def\ymin{-0.1}
\def\ymax{0.75}
\centering
  \begin{tikzpicture}
    \def\paramMargin{0.05}
    \def\paramDelta{0.1}
  \pgfmathsetmacro{\deltaxRight}{(1+\paramDelta)}
  \pgfmathsetmacro{\deltaxLeft}{1/(1+\paramDelta)}
  \begin{axis}[
  	height=\height,
  	width=\width,
    %	samples=200, % can be reduced when taking the straight lines below
    xmin=\xmin,
  	xmax=\xmax,
  	ymin=\ymin,
  	ymax=\ymax,
    extra x ticks={\deltaxLeft,\deltaxRight},
    extra x tick labels={$\frac{1}{1+\delta}$,$1+\delta$},
    extra x tick style={xticklabel style={yshift=-8, font=\footnotesize}}
    %xlabel=$\frac{R_{1,i}}{R_{2,i}}$,
    %ylabel={$\ell_{\delta}(J, R)_{\text{Hinge}}$}
  ]
    % \addplot+[red,mark=none]{max(0, x - (1/(1+\paramDelta+\paramMargin)))};
    \addplot [red, line width=\thickness] coordinates { (-2, 0)  (1/(1+\paramDelta+\paramMargin), 0) (1/(1+\paramDelta+\paramMargin)+4, 4) };
    % \addplot+[green,mark=none]{max(0, 1/(1+\paramDelta-\paramMargin) - x, x - (1+\paramDelta-\paramMargin)};
    \addplot [darkgreen, line width=\thickness] coordinates { (1/(1+\paramDelta-\paramMargin)-2, 2) (1/(1+\paramDelta-\paramMargin), 0) (1+\paramDelta-\paramMargin, 0) (1+\paramDelta-\paramMargin+4, 4) };
    % \addplot+[blue,mark=none]{max(0, (1+\paramDelta+\paramMargin) - x)};
    \addplot [blue, line width=\thickness] coordinates { (1+\paramDelta+\paramMargin-2, 2) (1+\paramDelta+\paramMargin, 0) (4, 0) };
    % show decision boundaries
    \addplot [dashed, line width=1] coordinates { (\deltaxLeft,\ymin) (\deltaxLeft,\ymax) };
    \addplot [dashed, line width=1] coordinates { (\deltaxRight,\ymin) (\deltaxRight,\ymax) };
  \end{axis}
  % label the ground truth decisions (I cannot find a quick solution to have it inside the axis environment above, therefore manually tuned position)
  \node [red]       at (1.4, 1.8) {$J_i=2$};
  \node [darkgreen] at (3.0, 1.8) {$J_i=E$};
  \node [blue]      at (4.6, 1.8) {$J_i=1$};
\end{tikzpicture}
%}
\caption{Visualization of the \WHDRhinge~loss dependent on the ratio ${R_{i_1}}/{R_{i_2}}$ for $\delta=0.1$ and $\margin=0.05$. The value of $\delta$ controls where the decision boundary for darker/lighter or equal reflectance lightness is made. With the value $\margin$, a margin from this boundary is encouraged. For values $\margin>\delta$ the $E$ class will always have a non-zero loss.
%\tn{either reflectance or lightness?!} \pg{do not understand} \tn{one of the two should be enough in the caption: ``darker/lighter or equal reflectance lightness''}
%\tn{Only show values above 1 and say otherwise exchange meaning of 1 and 2.}
}
\label{fig:whdr_hinge}
\end{figure}

\endgroup

%% with $\ell_\delta$ being the weighted 0/1 loss  % chktex 8
%% % \begin{align}
%% % \ell_\delta\left(J_i,\frac{R_{1,i}}{R_{2,i}}\right) = \mathbb{1}\left(J_i\neq\hat J_{i,\delta}(R)\right),
%% % \label{eq:0-1-loss}
%% % \end{align}
%% \begin{align}
%% \ell_\delta\left(J,R,i\right) = \mathbb{1}\left(J_i\neq\hat J_{i,\delta}(R)\right).
%% \label{eq:0-1-loss}
%% \end{align}
%% where $J_i$ is the human judgment (considered as ground truth), given as $1$ if the lightness $R_{i,1}$ of point 1 of the comparison is darker than $R_{2,i}$, that of point 2. It is 2 if the lightness of point 2 is darker, and E if they are more or less the same.

% \tn{probably omit next equation}
% For ease of understanding the following, we simply rewrite (\ref{eq:judgmentRewritten}) without any semantic change:
% \begin{align}
% \hat J_{i,\delta}(R) = \begin{cases}
% 1 & \text{if } \frac{R_{1,i}}{R_{2,i}} < \frac{1}{1+\delta}\\
% E & \text{if } \frac{1}{1+\delta} \leq \frac{R_{1,i}}{R_{2,i}} \leq 1+\delta\\ % chktex 21
% 2 & \text{if } 1+\delta < \frac{R_{1,i}}{R_{2,i}}
% \end{cases}
% \end{align}

We construct a proxy loss for the WHDR that can be used for supervised training.
The formulation is an adaption of the $\varepsilon$-insensitive loss for regression~\cite{vapniknature} for this problem setup.
We define
%\begin{align}
%\ell_{\delta,\margin}\left(J,R,i\right) &= \begin{cases}
%\max\left(0,\ \frac{R_{p_i}}{R_{q_i}} - \frac{1}{1+\delta+\margin}\right) &\text{if }J_i=1\\
%\max\left(0,\ \frac{1}{1+\delta-\margin}-\frac{R_{p_i}}{R_{q_i}},\ \frac{R_{p_i}}{R_{q_i}}-(1+\delta-\margin)\right) & \text{if}\ J_i=E\\
%\max\left(0,\ 1+\delta+\margin - \frac{R_{p_i}}{R_{q_i}}\right) & \text{if}\ J_i=2,
%\end{cases}\label{eq:lR}
%\end{align}
\begin{align}
&\ell_{\delta,\margin}\left(J,R,i\right) =
\notag\\
&\begin{cases}
\max\left(0,\ \frac{R_{i_1}}{R_{i_2}} - \frac{1}{1+\delta+\margin}\right) &\text{if }J_i=1\\
\max\left(0,\begin{cases}
\frac{1}{1+\delta-\margin}-\frac{R_{i_1}}{R_{i_2}},\\
\frac{R_{i_1}}{R_{i_2}}-(1+\delta-\margin)
\end{cases}\right)& \text{if}\ J_i=E\\
\max\left(0,\ 1+\delta+\margin - \frac{R_{i_1}}{R_{i_2}}\right) & \text{if}\ J_i=2,
\end{cases}\label{eq:lR}
\end{align}
which is visualized
%for two different values of $(\delta,\margin)$
in~\cref{fig:whdr_hinge}.
The scalar $\delta$ is the threshold of the WDHR$_{\delta}$ and we introduce the hyper-parameter~$\margin$, which is the margin between the neighbouring classes $1,E$ and~$2,E$.

The pipeline of supervised training is simple.
A network produces a dense decomposition $R$, which is then used to arrive at relative judgments for two pixel locations based on the ratio of the predicted $R$ values.
The loss in~\cref{eq:lR} is then weighted and summed over all annotated pixel pairs, similar to~\cref{eq:whdr}, and the error is propagated backwards to compute the gradients of the network parameters.

As with the standard hinge-loss commonly used for binary SVM training, the sub-gradients of the WHDR-hinge loss can be easily computed.

\subsection{Train and test data set}
The IIW dataset does not come with a pre-defined train, validation and test split.
We adopt the split suggested by~\cite{narihira2015lightness} into $80\%$ training and $20\%$ test images, putting the first of every five images sorted by file name in the test set.
In order to properly evaluate different models, we additionally split the data into a separate validation set, with the ratios of $70\%$ training, $10\%$ validation and $20\%$ test.
We keep the test set of~\cite{narihira2015lightness}, and use from every series of $10$ images the seventh in the validation set.

%%%%%%%%%%%%%%%%%%%%
\subsection{Network architecture of the CNN}
We take the linearized RGB images in the range $[0,1]$ as input, evaluate a series of $n$ convolutional layers with~$f$~filters each, acting on a kernel of size~$k$, with a ReLU as nonlinear activation function in between. The padding in the convolutions is chosen based on $k$, so as to not change the resolution. The output of all nonlinearities is concatenated and convolved with a $1\times1$ filter to fuse the information of skipped layers. A last sigmoidal activation function bounds the single channel output~$r$, on which the \WHDRhinge{} loss, as given in~\cref{sec:whdr_hinge}, operates during training.

One final layer recovers RGB reflectance~$R$ and shading~$S$ from the scalar reflectance intensity~$r$, as given in~\cref{eq:recover_reflectance_shading}, to output the final dense intrinsic image decomposition.

%%%%%%%%%
\paragraph{Resolving light intensity}%\label{sec:boundaries}
The last nonlinearity in the network acting on $r$ is included since ambiguity about the light intensity in an image cannot be solved. It is only possible to determine reflectance and shading up to a constant $\alpha\in (0,\infty)$, since $I=RS=\left(\alpha R\right)\left(\frac{1}{\alpha}S\right)$.
%Therefore, we include a second term $\ell(t) = \max(0,-t,t-1)$ to the objective in order to penalize values of $S$ that fall outside the $[0,1]$ interval.
Therefore, to keep the reflectance values bounded, we employ a sigmoidal activation function to limit the scalar reflectance intensity to be in the range $[0,1]$.
%This will automatically adjust for the scalar ambiguity and resolve the over-parameterization of the decomposition.
%The final loss is the addition of the \WHDRhinge{} on $R$ and this term on $S$ scaled by~$10^{-2}$.

%%%%%%%%%%%%%%%%%%%%
\subsection{Experiments}\label{sec:cnn_experiments}
For all experiments in this paper we use the open source deep learning framework caffe~\citep{jia2014caffe} utilizing the ADAM solver~\cite{kingma2015adam} with a learning rate of $0.001$, momentum of $\beta_1=0.9$ and momentum-2 of $\beta_2=0.999$. All training images are resized to a fixed $256\times256$ pixel resolution to be able to process them in batches.
More details about data augmentation and label analysis are included in the supplementary.

\begin{figure}[t]
\centering
\vspace{-8pt}
\subfigure[]{
\includegraphics[width=0.43\linewidth]{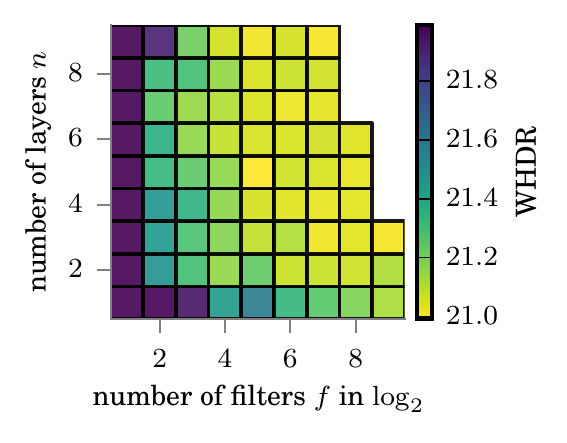}
\label{fig:numLayers_numFilters}
}
\hspace{-0.075\linewidth}
\subfigure[]{
\includegraphics[width=0.58\linewidth]{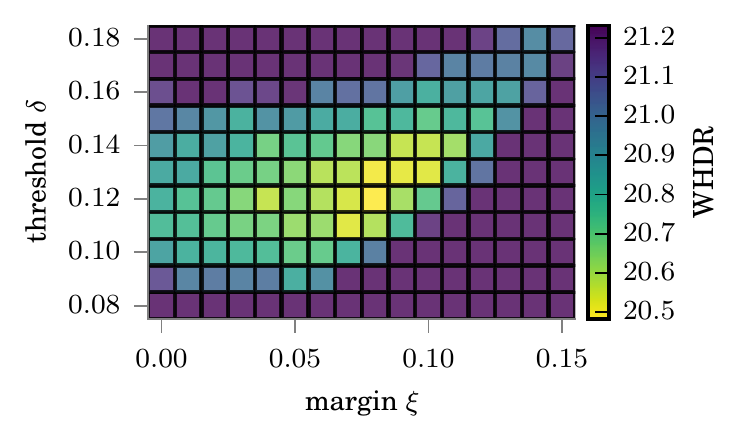}
\label{fig:whdr_delta_margin}
}
\caption{
\subref{fig:numLayers_numFilters} WHDR for different network depths~$n$ and number of filters~$f$. Missing data is the result of memory limit on our graphics card.
\subref{fig:whdr_delta_margin} WHDR for different thresholds~$\delta$ and margins~$\margin$.
}
\end{figure}

% Network Hyper-Parameters
%%%%%%%%%%
\paragraph{Network Hyper-Parameters} For the network layout described above we performed an extensive parameter sweep over a varying number of kernel widths~$k\in\{1,3,5,7\}$, layers~$n\in\{1,\ldots,9\}$, and filters~$f\in\{2^1,\ldots,2^9\}$.
The results on the validation set for $k=1$ are shown in~\cref{fig:numLayers_numFilters}.
The number of layers $n$ has only small influence on the performance above $n\geq2$, similar with $f\geq2^4$.
An unexpected finding was that the kernel size $k$ has little to no effect and $1\times1$ convolutions work just as well as those with bigger kernels. This means that the network only learns a pixel-wise lookup table, but at the same time, the network performs already better in WHDR than most state-of-the-art methods in the literature.
This amounts to a re-scaling of the reflectance intensity at every pixel separately, no context needed. We will discuss this further in~\cref{sec:networkDiscussion}.
From this analysis we chose a network of $n=5,f=2^5$, and $k=1$ as the basis for all future experiments.

In addition to this basic setup we also played with different network layouts, \eg, without skip connections and with a U-net like architecture~\cite{ronneberger2015uNet}, tried PReLU~\cite{he2015prelu} nonlinearities in between and dilated convolutions~\cite{yu2015dilation} to widen the receptive field, but did not find better results.
In general we found that simpler networks perform better, what we believe is the outcome of the amount of weakly labeled training data.
%We note that existing works on low-level vision tasks (\eg~\cite{}) also observed similar behavior, indicating that deeper networks do not necessarily result in better performance. %\tn{find citation for [Dong et al. TPAMI-15]}

%%%%%%%%%%%%%%%%%%%%%%%
% two rows
%%%%%%%%%%%%%%%%%%%%%%%
%\begin{figure}[t]
%\begin{center}
%\subfigure[]{
%\includegraphics[width=0.8\linewidth]{numLayers_numFilters_landscape_1028_kernels_1x1_skip_layers_batchsize2}
%\label{fig:numLayers_numFilters}
%}
%%\hfill
%\subfigure[]{
%% \includegraphics[width=\linewidth]{whdr_delta_margin_after_full_10000_landscape}
%% \includegraphics[width=0.49\textwidth]{whdr_delta_margin_best_in_10000_landscape}
%\includegraphics[width=\linewidth]{whdr_delta_margin_best_in_all_landscape}
%\label{fig:whdr_delta_margin}
%}
%\end{center}
%\caption{
%\subref{fig:numLayers_numFilters} WHDR for different network depths~$n$ and number of filters~$f$. Missing data is the result of memory limit on our graphics card.
%\subref{fig:whdr_delta_margin} WHDR for different thresholds~$\delta$ and margins~$\margin$.
%}
%\end{figure}
%%%%%%%%%%%%%%%%%%%%%%%%%%%%%
% one row
%%%%%%%%%%%%%%%%%%%%%%%%%%%%%

% Threshold and margin parameters of the WHDR hinge loss
%%%%%%%%%%
\paragraph{Hyperparameters of the \WHDRhinge~loss}
To minimize the WHDR rate consistent with~$\delta=0.1$ from~\cite{bell2014iiw} we optimized the loss hyper-parameter $\delta,\margin$ on the validation set. The influence is shown in~\cref{fig:whdr_delta_margin} and the final parameters used for training are $\delta=0.12$ and $\margin=0.08$.

%%%%%%%%%%%%%%%%%%%%
\subsection{Discussion of the results}\label{sec:networkDiscussion}
\begin{figure}[t]
%% tiled RGB
%\includegraphics[width=0.49\linewidth]{lookup_table/lookup_table_all-input}
%\includegraphics[width=0.49\linewidth]{lookup_table/lookup_table_all-r}
% CNN output directly for varying hue and saturation in HSV for value=255
\includegraphics[width=0.49\linewidth]{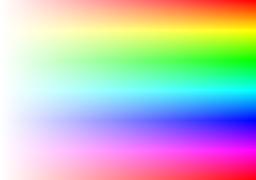}
\includegraphics[width=0.49\linewidth]{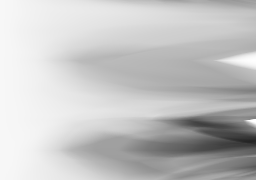}
\caption{
Lookup table in HSV space, generated by our direct prediction network, for varying hue and saturation and a constant value/brightness of $255$.
Left: The input image $I$, Right: The single channel reflectance intensity $r$ predicted by the CNN.
%\tn{Check the latex code to see  more possibilities or ask me if you want to see saturation or value for other values.}
}
\label{fig:lookup_table}
\end{figure}

Many methods build on the Retinex assumption~\cite{land1971retinex}, which states that strong image gradients are reflectance edges and small gradients are explained by shading.
Under the assumption of smooth shading, local gradient estimation would only require a small receptive field, but there is no possibility that a method can resolve shading from a single pixel alone, e.g., see the famous illusion of~\cite{adelson2000illusions} for a counter example.
Still, in terms of WHDR, this method performs better than most methods~\cite{shen2011sparse,garces2012clustering,zhao2012nonlocalTextureConstraints,bell2014iiw,zhou2015learningPriors} on IIW, see the table~\cref{tab:whdr} for an empirical comparison to a few approaches.
\begin{table}
\caption{Comparison of some intrinsic image approaches on IIW. An extended comparison is in~\cref{fig:whdr_boxplot}.
}
\label{tab:whdr}
\vspace{3pt}
\centering
\begin{tabular}{c|cccccc}
Method & Retinex & \cite{bell2014iiw} & \cite{zhou2015learningPriors} & \textbf{ours} & \cite{zoran2015ordinal} & \cite{bi2015l1} \\
WHDR   & $26.9$  & $20.6$             & $19.9$                        & $19.5$        & $17.9$                  & $17.7$
\end{tabular}
\end{table}
Since there is a direct pixel-to-reflectance relationship, we can visualize a ``lookup-table'' mapping RGB pixels to reflectance, see~\cref{fig:lookup_table}.
A big portion of colors is judged to have more or less the same reflectance intensity as white.
Blue is mostly judged being darker than green, which is biologically plausible. Green light contributes the most to the intensity perceived by humans, and blue light the least~\cite{poynton2012digitalVideo}.
There is a portion of very light reflectance for fully saturated green, even brighter than from white pixels. This may be a result of the Helmholtz-Kohlrausch effect~\cite{corney2009brightnessOfColour}, humans perceive colored light brighter than white light. This may lead to wrong human reflectance judgments under the circumstances of bright saturated colors.
%Either this finding suggests that there is a statistical bias in IIW, or it provides a lower bar that any reasonable intrinsic image decomposition should attain.

%%%%%%%%%%%%%%%%%%%%

%%% as a big lookup table

\begin{figure}[t]
\centering
%\vspace{-3pt}  % easy fix if we need a line more in this column
\subfigure[]{
\includegraphics[width=0.32\linewidth]{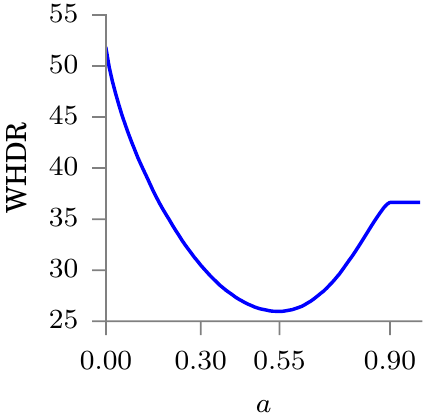}
\label{fig:lower_bound_scaling:plot}
}
\subfigure[]{
\includegraphics[width=0.2\linewidth]{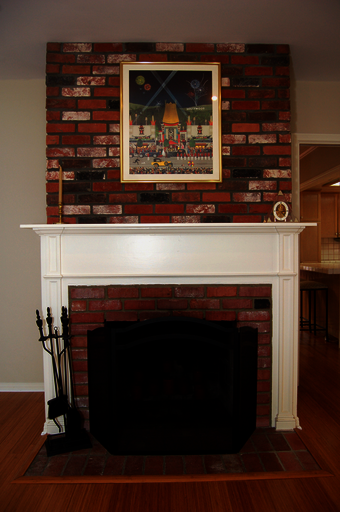}
\label{fig:lower_bound_scaling:input}
}
\subfigure[]{
\includegraphics[width=0.2\linewidth]{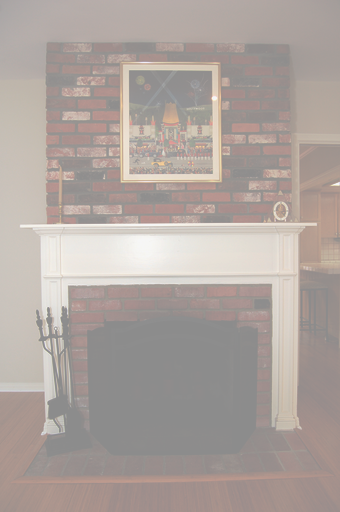}
\label{fig:lower_bound_scaling:scaled}
}
\caption{
\subref{fig:lower_bound_scaling:plot} Mean WHDR (in \%) on training and validation set. As reflectance image we take the input image after rescaling it into the range $[a,1]$.
\subref{fig:lower_bound_scaling:input} An example input image.
\subref{fig:lower_bound_scaling:scaled} The same image scaled to have a lower bound of $a=0.55$, where the mean WHDR is minimal. %In our opinion both should have equal WHDR.
}
\label{fig:lower_bound_scaling}
\end{figure}

%This result highlights a property of the WHDR loss.
A simple re-scaling of an image into $[a,1]$ with $a=0.55$  (see~\cref{fig:lower_bound_scaling} for an example), then using the image as a reflectance estimate
%(constant shading)
results to a WHDR score of $25.7$ on the test set.
This low score is due to an in-balance of relative judgments, $2/3$ of which are equal judgments.
A re-scaling to $[0.55,1]$ makes most equal judgments correct and compromises the unequal judgments.
The output of the CNN is on average in the range $[0.48,0.96]$ but its non-linearity accounts for small variations in color and therefore makes more un-equal judgements correct.
Remember that the CNN predicts a dense reflectance map on the test set, unaware of the point pairs performance will be evaluated on.
The CNN implements no explicit prior knowledge, we will show next that encouraging piecewise constant reflectance improves the result.

%%%%%%%%%%%%%%%%%%%%%%%%%%%%%%%%%%%%%%%%%%%%%%%%%%%%%%%%%%%%%%%%%%
\section{Reflectance Filtering}\label{sec:filtering}
A common prior of intrinsic image estimation is to have only a sparse set of reflectances present in a scene~\cite{omer2004colorlines,gehler2011sparsity,shen2011sparse}.
We describe a new technique to include this prior knowledge using an image filter.
This allows an easy integration into existing techniques and we found that filtering reflectance estimates always improves performance.

%%%%%%%%%%%%%%%%%%%%
\subsection{Image-aware filtering}
A general linear translation-variant filtering process is defined as
\begin{align}
q_i = \sum_j W_{ij}(I)p_j, \label{eq:general_filter}
\end{align}
where the input image $p$ is smoothed under the guidance of an image $I$ to the filtered output image $q$. Here $i,j$ denote pixels and the sum runs over the entire image. Two examples are the joint bilateral filter and the guided filter, whose weights for~\cref{eq:general_filter} we summarize next.

%%%%%%%%%%
\paragraph{The (joint) bilateral filter}
The joint bilateral filter~\cite{petschnigg2004jointBilateral} is an extension to the bilateral filter~\cite{tomasi1998bilateral} which uses feature difference in a (potentially different) guidance image to spatially smooth pixels in the input image. It defines the weights as
\begin{align}
W_{ij}(I) = \frac{1}{K_i}\exp\left(-\frac{\vert x_i - x_j\vert^2}{\sigma_s^2}-\frac{\vert I_i - I_j\vert^2}{\sigma_r^2}\right),
\label{eq:bilateral_filter}
\end{align}
with $x_i$ being pixel coordinates.
This means that pixels that are both close spatially and in intensity in the guidance image will be smoothed more. The normalization $K_i$ is chosen to ensure $\sum_j W_{ij}=1$.

%%%%%%%%%%
\paragraph{The guided filter}
The guided filter~\cite{he2010guided} is a fast alternative to the joint bilateral filter, it is also edge-preserving, and has better behavior near edges. It is based on a locally linear model
$\forall i\in\omega_k: q_i = a_kI_i + b_k$,
where $a_k$, $b_k$ are linear coefficients assumed to be constant in the square window $\omega_k$ centered at pixel $k$ of size $r$.
The linearity guarantees that $q$ has an edge only if $I$ has an edge, since $\nabla q = a\nabla I$.
Solving for the coefficients that minimize the difference between $q$ and $p$ leads to the weights
\begin{align}
W_{ij}(I) = \frac{1}{\vert\omega\vert^2}\sum_{k:(i,j)\in\omega_k}\left(1+\frac{(I_i-\mu_k)(I_j-\mu_k)}{\sigma_k^2+\varepsilon}\right)
\label{eq:guided_filter}
\end{align}
where $\mu_k$ and $\sigma_k^2$ are the mean and variance of $I$ in $\omega_k$, $\vert\omega_k\vert$ is the number of pixels in $\omega_k$ and $\varepsilon$ a constant parameter similar to the range variance $\sigma_r^2$ in the bilateral filter.
%It can be shown that $\sum_j W_{ij}(I)=1$, so no further normalization is needed.
Especially for larger spatial scales, the guided filter benefits from not having the quadratic dependency on the filtering kernel size. We refer to~\cite{he2010guided} for a more thorough discussion.

%%%%%%%%%%%%%%%%%%%%
\subsection{Filtering for piecewise constancy}\label{sec:filtering_results}
We need to define a guidance image to fully specify the filtering operation.
An ideal guidance image would group pixels into regions of constant reflectance.
We will refer the filtered image with \mbox{BF(\emph{method}, \emph{guidance})} for the bilateral filter and \mbox{GF(\emph{method}, \emph{guidance})} for the guided filter, respectively.

%%%%%%%%%%%%%%%%%%%%
\paragraph{Using a flattened image as guidance}
The method of~\cite{bi2015l1} formulates an optimization problem to group pixels into regions of similar reflectance. This provides a good candidate for a suitable guidance image.
The piecewise flattened image is found by minimizing
%\begin{align}
$E = E_l + \alpha E_g + \beta E_a$,
%\end{align}
with the \emph{local flattening energy}
\begin{align}
E_l = \sum_i\sum_{j\in N_h(i)}\exp\left(-\frac{\Vert f_i - f_j\Vert_2^2}{2\sigma^2}\right)\Vert q_i-q_j\Vert_1,
\label{eq:bi:local_flattening}
\end{align}
where $N_h(i)$ is the $h\times h$ neighborhood of the $i$-th pixel, $q_i$ is the output RGB vector, $f_i=[\kappa\cdot l_i,a_i,b_i]$, with $[l_i,a_i,b_i]$ being the input vector in CIELab color space and $\kappa,\sigma$ are hyper-parameters.
A \emph{global sparsity energy} is defined as
\begin{align}
E_g = \sum_{i\in S_r}\sum_{j\in S_r} w_{ij}\Vert q_i-q_j\Vert_1,
\end{align}
with the same affinity weights $w_{ij}$ as in~\cref{eq:bi:local_flattening} and $S_r$ being the set of representative pixels which are closest to the average color in their superpixels.
To avoid the trivial solution, a \emph{data term for image approximation} is added:
\begin{align}
E_a = \Vert q - p\Vert_2^2.
\end{align}
See~\cite{bi2015l1} for how to solve the resulting optimization problem.
%We will refer to the result of this operation as the \emph{flat} guidance image.
We will refer to the result of this $L_1$-flattening optimization from now on also simply as \emph{flat}.

\paragraph{Filtering Results}
Using ``flat'' as guidance, we again found the best hyper-parameters on the validation set ($\sigma_r=15,\sigma_s=28$ for the joint bilateral and $r=45,\varepsilon=3$ for the guided filter).
Using the CNN predictions as input to the filter, we find the result of BF(CNN,flat) to improve to $18.1$ and GF(CNN,flat) further to $17.7$.
This is on par with the current state-of-the art ($17.67$) which is the full pipeline of~\cite{bi2015l1}: the flat image is clustered, followed by a CRF and another energy minimization step. 
We note that using the $L_1$ flattened result directly as reflectance image has $20.9$ WHDR, which shows that there is complementary information in the CNN output and the guidance image.

This use of the flattened image as a guidance in a filter, extends~\cite{bi2015l1} and allows application to other intrinsic image decompositions.
We apply filtering to the second best method~\cite{zoran2015ordinal} on IIW.
Their work proposes to create a sparse representation of the image by using the centers of a superpixelization.
Patches around those centers are extracted and a CNN is used to provide an ordinal relationship via the three-way classification into ``darker'', ``equal'', and ``lighter''.
This sparse result is then again densified by solving a constrained quadratic optimization problem to produce a full reflectance image.

The application of GF(\cite{zoran2015ordinal}, flat) improves WHDR from $17.85$ to $16.38$.
Repeated application of the filter further improves the output down to $15.78$ after three applications of the guided filter.
This result represents the new state-of-the-art by a large margin.

%%%%%%%%%%%%%%%%%%%%%%%%%%%%%%%%%%%%%%%%%
% Results of filtering several times on validation set
%      median, mean
%%%%%%%%%%%%%%%%%%%%%%%%%%%%%%%%%%%%%%%
% ours: 19.91, 20.55
% ours guided with ours c52s7:
% 1x    19.33, 20.17
% 2x    19.57, 20.20  % already worse
% 3x    19.87, 20.27  % even worse

% ours bilateral with ours: c20s22, runtime: 1.3 - 2.5 s/it on 8 CPUs
% 1x    18.91, 19.73
% 2x    19.24, 19.69  % plateaued (see median))
% 3x    18.91, 19.75

% ours guided with flat c3s45:
% 1x    17.62, 18.44
% 2x    17.73, 18.47  % already worse
% 3x    17.87, 18.56  % even worse

% ours bilateral with flat c20s22, runtime: about 5-6 s/it on 4 CPUs
% 1x    18.75, 19.14
% 2x    18.19, 18.92
% 3x    17.79, 18.87
% 4x    17.92, 18.87 % slightly worse, did not reach guided with one step

%%%%%%%%%%%%%%%%%%%%%%%%%%%%%%%%%%%%%
% Bi:   17.51, 18.23
% guided with flat c3s45:
% 1x    17.29, 18.43 % already worse
% 2x    17.39, 18.63
% 3x    17.54, 18.82

% bilateral with flat: c20s22
% 1x    17.12, 18.14
% 2x    17.13, 18.13
% 3x    17.18, 18.14  % reached best probably already with one step, works way better than guided

%%%%%%%%%%%%%%%%%%%%%%%%%%%%%%%%%%%%
% Zoran:
% Proper evaluation of filtering is problematic since we only have Zoran's test set, no validation set
%%%%%%%%%%%%%%%%%%%%%%%%%%%%%%%%%%%%

\newcolumntype{M}{>{\centering\let\newline\\\arraybackslash\hspace{0pt}}m{20.9mm}}
\newcommand{\mcA}[1]{\multicolumn{1}{M}{#1}}
\newcolumntype{N}{>{\centering\let\newline\\\arraybackslash\hspace{0pt}}m{17.9mm}}
\newcommand{\mcB}[1]{\multicolumn{1}{N}{#1}}
\newcolumntype{O}{>{\centering\let\newline\\\arraybackslash\hspace{0pt}}m{18.1mm}}
\newcommand{\mcC}[1]{\multicolumn{1}{O}{#1}}
\begin{table}[t]
\caption{Comparison of filtering performance for intrinsic image estimation methods under varying guidance images.
We report the improvement in mean WHDR over the images in the test split from Narihira et al.~\cite{narihira2015lightness}, and for~\cite{zoran2015ordinal} results on their respective test set (marked with an asterisk) before and relatively to it after one filtering operation.}
\label{tab:filter_comparison}
\centering
%\begin{center}
\vspace{5pt}  % somehow there was really not much space between caption and table
\begin{tabular}{c|S[table-format=2.2]S[table-format=2.2]S[table-format=2.2]}
Method     & \mcA{F(CNN, CNN)} & \mcB{F(CNN, flat)} & \mcC{F(\cite{zoran2015ordinal}, flat)*} \\
\hline
unfiltered & 19.49             & 19.49                 & 17.85                                 \\
\hline
%bilateral  & 18.89             & 18.11                 & 16.38                                 \\
BF  & -0.6              & -1.38                 & -1.47                                 \\
%guided     & 19.24             & 17.69                 & 15.87                                 \\
GF     & -0.25             & -1.8                  & -1.98
\end{tabular}
%\end{center}
% if we want Bi final with Bi flat (on test): unfiltered 17.67, bilateral 17.46 (-0.41), guided 17.87 (+0.2),
\end{table}

\begin{figure}[t]
\centering
\includegraphics[width=\linewidth]{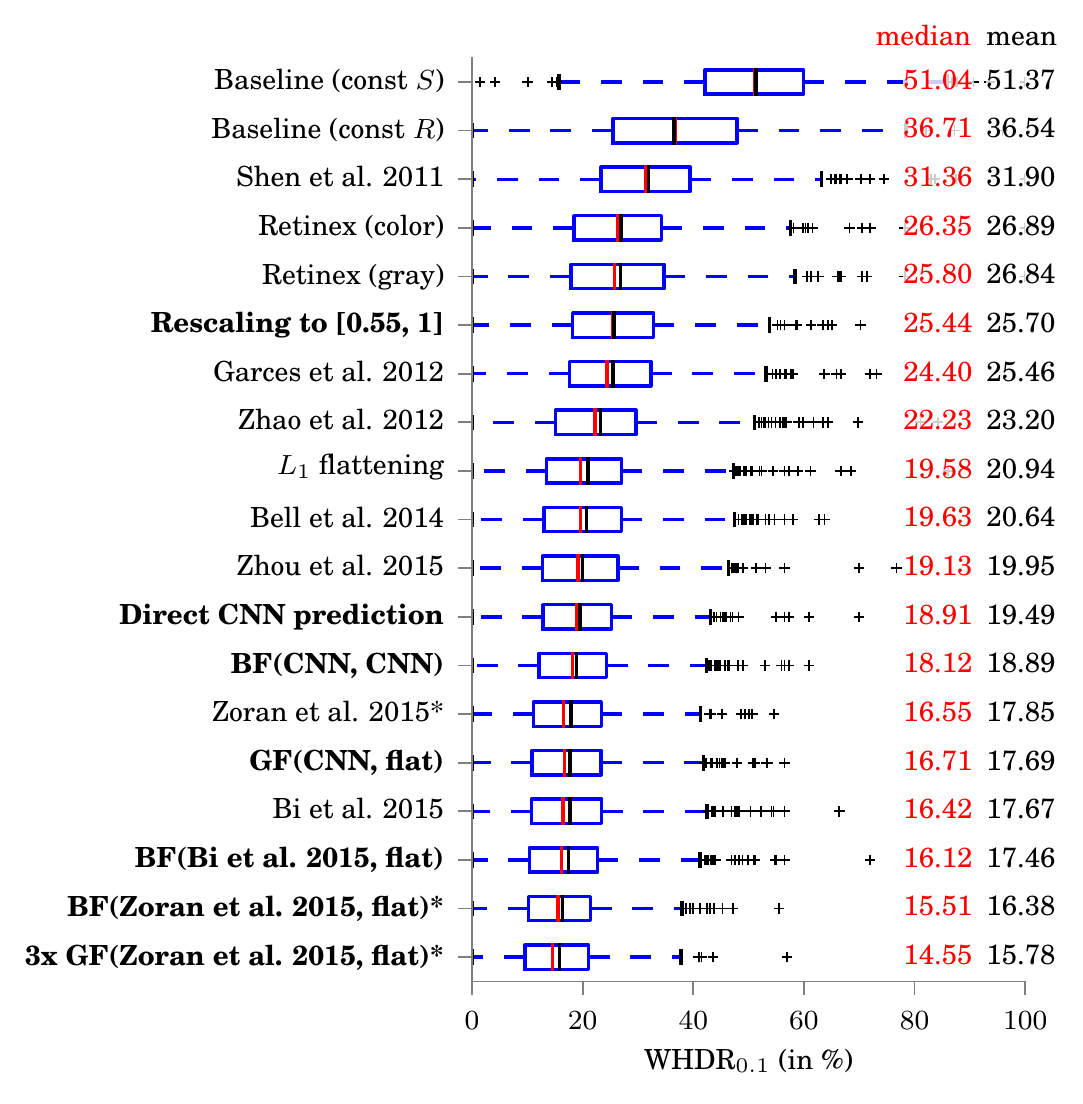}
\caption{Comparison of performance to different intrinsic image estimation methods. Over all images in the test split from Narihira et al.~\cite{narihira2015lightness}, we report the statistics of the individual WHDRs on the images. The red line represents the median, the black line the mean.
Results of Zoran et al. 2015~\cite{zoran2015ordinal}, are based on reflectance predictions provided by the authors which are generated on a different test split. All methods that are evaluated on this different test set are marked with an asterisk as they are not directly comparable.
%The results using Zoran et al. 2015~\cite{zoran2015ordinal}, are evaluated on a different test set, therefore marked with an asterisk, since not directly comparable.
}
\label{fig:whdr_boxplot}
\end{figure}

\paragraph{Filtering with the image itself}
A conceptually easier choice is to filter using the input image itself as guidance.
We applied this to the Direct CNN predictions and searched for the hyper-parameters (see supplementary) to find $\sigma_s=22, \sigma_r=20$. Bilateral filtering with the input already improved test performance from $19.5$ to $18.9$. Guided filtering also improved a bit to $19.2$ with $r=7$ and $\varepsilon=52$.

\subsection{Discussion}
We found throughout that guided filtering improves performance, see~\cref{tab:filter_comparison}.
There are some cases where the joint bilateral filter outperforms the guided filter, but in general the latter has better performance.
Also the guided filter is magnitudes faster.
We summarize a comparison with recent methods and state-of-the-art in~\cref{fig:whdr_boxplot}.
% Please also see the supplemental material for a more thorough comparison of results from the joint bilateral and guided filter.

As expected, the quantitative performance increases also the qualitative results.
For an impression to assess the qualitative performance, we refer to~\cref{fig:qualitative}.
Results on a larger number of sample images and in comparison to more related work can be found in the supplementary material.

\def\fileIDa{71341}
\def\fileIDb{58346}
\def\size{0.235\linewidth}
\def\shortenSpace{-6pt}

\begin{figure*}[t]
\centering
\begin{minipage}{\size}
\subfigure[input image]{
\includegraphics[width=\linewidth]{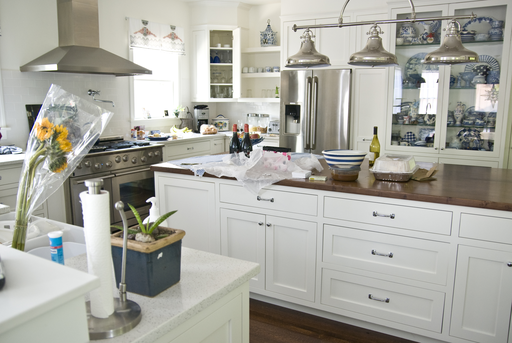}
\label{fig:qualitative:a:input}
}\\\vspace{\shortenSpace}
\subfigure[reflectance of~\cite{zoran2015ordinal}]{
\includegraphics[width=\linewidth]{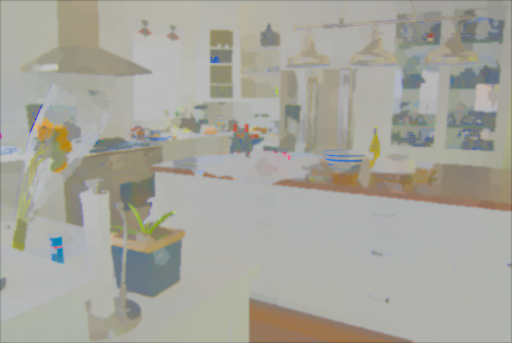}
\label{fig:qualitative:a:zoran-r}
}\\\vspace{\shortenSpace}
\subfigure[shading of~\cite{zoran2015ordinal}]{
\includegraphics[width=\linewidth]{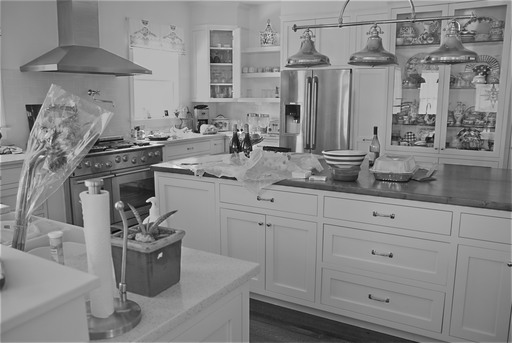}
\label{fig:qualitative:a:zoran-s}
}
\end{minipage}
\begin{minipage}{\size}
\subfigure[flat guidance]{
\includegraphics[width=\linewidth]{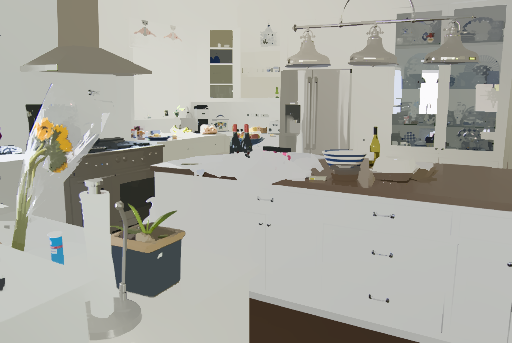}
\label{fig:qualitative:a:guidance}
}\\\vspace{\shortenSpace}
\subfigure[filtered reflectance]{
\includegraphics[width=\linewidth]{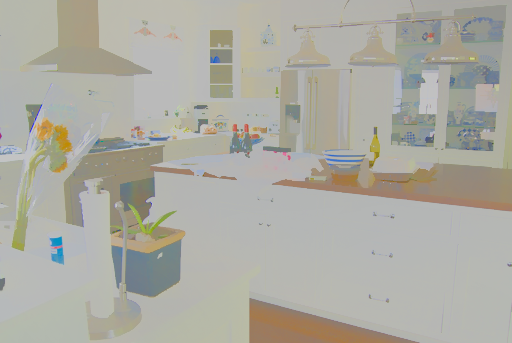}
\label{fig:qualitative:a:filtered-r}
}\\\vspace{\shortenSpace}
\subfigure[respective shading]{
\includegraphics[width=\linewidth]{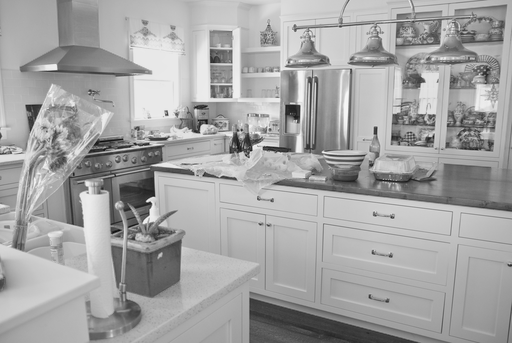}
\label{fig:qualitative:a:filtered-s}
}
\end{minipage}
\hfill
\begin{minipage}{\size}
\subfigure[input image]{
\includegraphics[width=\linewidth]{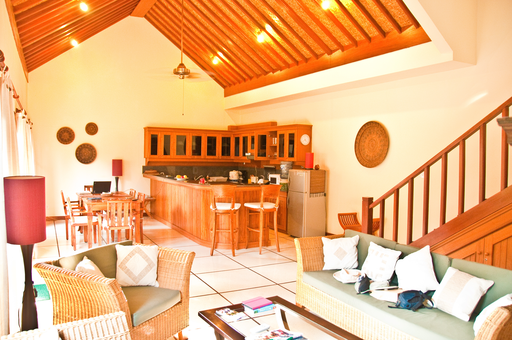}
\label{fig:qualitative:b:input}
}\\\vspace{\shortenSpace}
\subfigure[reflectance of~\cite{zoran2015ordinal}]{
\includegraphics[width=\linewidth]{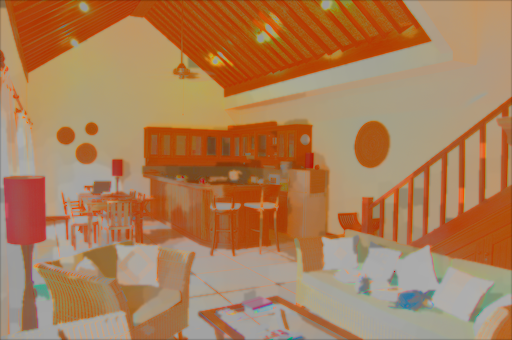}
\label{fig:qualitative:b:zoran-r}
}\\\vspace{\shortenSpace}
\subfigure[shading of~\cite{zoran2015ordinal}]{
\includegraphics[width=\linewidth]{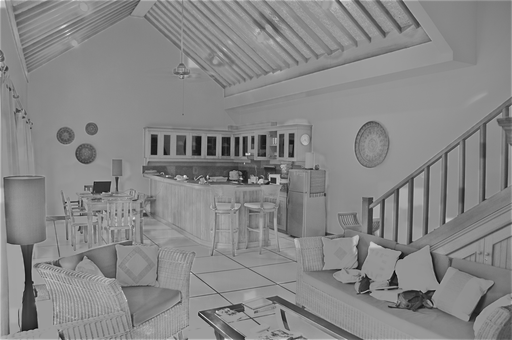}
\label{fig:qualitative:b:zoran-s}
}
\end{minipage}
\begin{minipage}{\size}
\subfigure[flat guidance]{
\includegraphics[width=\linewidth]{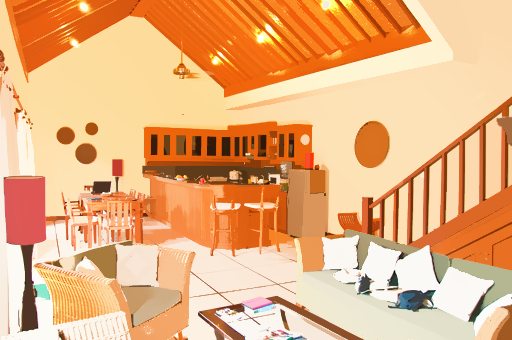}
\label{fig:qualitative:b:guidance}
}\\\vspace{\shortenSpace}
\subfigure[filtered reflectance]{
\includegraphics[width=\linewidth]{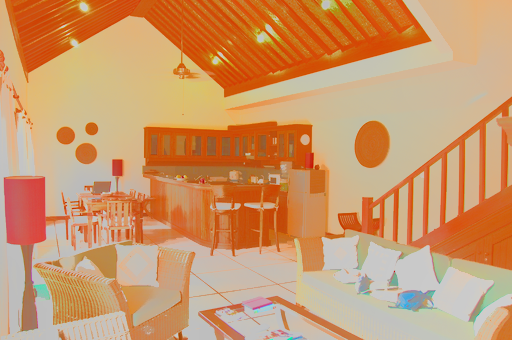}
\label{fig:qualitative:b:filtered-r}
}\\\vspace{\shortenSpace}
\subfigure[respective shading]{
\includegraphics[width=\linewidth]{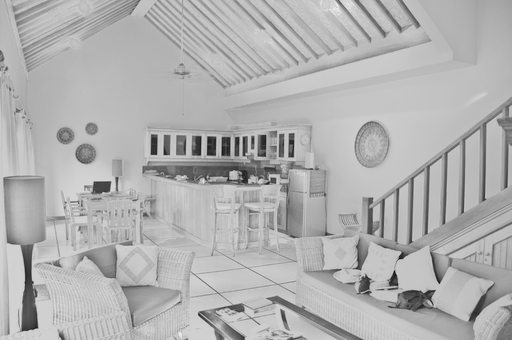}
\label{fig:qualitative:b:filtered-s}
}
\end{minipage}
\caption{Sample decompositions of the
\subref{fig:qualitative:a:input} input image with the IIW ID~\fileIDa{} into
\subref{fig:qualitative:a:zoran-r} reflectance and
\subref{fig:qualitative:a:zoran-s} shading by the method of~\cite{zoran2015ordinal}. Filtering it using the flat guidance image in
\subref{fig:qualitative:a:guidance} results in the intrinsic layers
\subref{fig:qualitative:a:filtered-r} and
\subref{fig:qualitative:a:filtered-s}
of our final model.
\subref{fig:qualitative:b:input}-\subref{fig:qualitative:b:filtered-s} are the same as~\subref{fig:qualitative:a:input}-\subref{fig:qualitative:a:filtered-s} for IIW ID~\fileIDb{}.
}
\label{fig:qualitative}
\end{figure*}

\begin{figure}[t]
\centering
\includegraphics[width=0.95\linewidth]{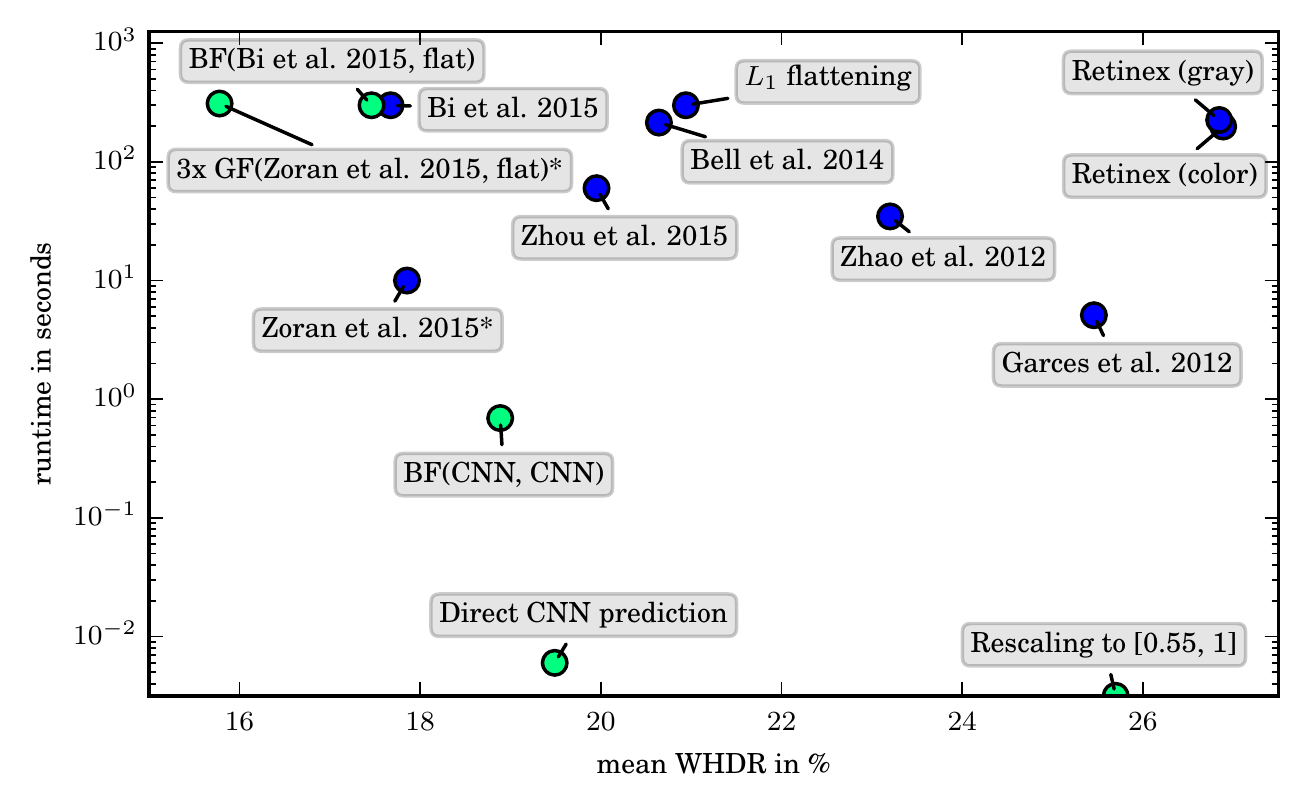}
\caption{WHDR against runtime for related work. Mean WHDR of competing methods is evaluated on the Narihira test split~\cite{narihira2015lightness} on the decompositions provided by~\cite{bell2014iiw} and the project pages of~\cite{bi2015l1,zhou2015learningPriors}. Methods with an appended asterisk were evaluated on the test split given in~\cite{zoran2015ordinal}. For methods which are evaluated in~\cite{bell2014iiw}, we used the reported runtimes on the corresponding project page.
Methods developed in this work are plotted in green, previous results are plotted in blue.
}
\label{fig:runtime_vs_whdr}
\end{figure}

\paragraph{Runtime analysis}
In~\cref{fig:runtime_vs_whdr} we show the runtime of different algorithms against their WHDR.
All methods of this paper are colored in green.
We collected the timing estimates from the respective statements in the corresponding publications.
By construction, our direct prediction CNN with only a few filters is fast at test time (\SI{180}{fps} on GPU) but it requires further filtering for better results.
The bilateral filter adds around \SI{2}{s} per image on CPU and the guided filter less than \SI{0.1}{s}.
The bottleneck of the filtering approach is the computation of the  $L_1$ flattened guidance image.

%%%%%%%%%%%%%%%%%%%%%%%%%%%%%%%%%%%%%%%%%%%%%%%%%%%%%%%%%%%%%%%%%%
\section{Conclusion}

In this paper we have proposed methods that are on opposing ends of employed prior knowledge.
This led to both the best results on IIW and valuable insights into the current state of intrinsic image estimation.
We presented the first end-to-end CNN method, trained on the \WHDRhinge~loss, that predicts a dense result without any post-processing step.
Our finding is that a context-free per-pixel judgment is sufficient for competitive results.
We believe that this should set a new lower bar for learning methods on~IIW.
While this observation may be attributed to an inherent bias in IIW, we have no qualified reason to believe so.
We still conjecture that good results correlate with low WHDR numbers, and note that human performance sets a high bar with a median WHDR of only $7.5\%$~\cite{bell2014iiw}.
This has not been attained by any automatic method so far.
We further develop a filtering technique to implement the assumption of piecewise constant and sparse reflectance.
This extends the work of~\cite{bi2015l1} and makes it possible to apply their reflectance grouping to other decompositions.
We find that a filtered CNN output is on par with the best published learning based methods and further improve the initial result of~\cite{zoran2015ordinal} to $15.78\%$
on its testset,
which is the lowest WHDR performance for a dense decomposition.

In summary, the findings of this paper suggest that it is still the use of strong prior knowledge in intrinsic estimation algorithms that drives empirical performance.
More research will be necessary to build combined models and enable learning from sparse pairwise judgments.
A future direction is to replace the expensive optimization of~\cite{bi2015l1} with CNN inference, which would enable fast algorithms for high quality intrinsic video decompositions.

% The code, models, and results that are necessary to re-produce the experiments of this paper are available at \url{https://ps.is.tue.mpg.de/research_projects/reflectance-filtering}.
All code, models, and results are available at \url{https://ps.is.tue.mpg.de/research_projects/reflectance-filtering}.

{\small
\bibliographystyle{ieee}
\bibliography{intrinsic}
}

%%%%%%%%%%%%%%%%%%%%%%%%%%%%%%%%%%%%%%%%%%%%%%%%%%%%%%%%%%%%%%%%%%
%%%%%%%%%%%%%%%%%%%%%%%%%%%%%%%%%%%%%%%%%%%%%%%%%%%%%%%%%%%%%%%%%%
%%%%%%%%%%%%%%%%%%%%%%%%%%%%%%%%%%%%%%%%%%%%%%%%%%%%%%%%%%%%%%%%%%
% SUPPLEMENTARY MATERIAL
%%%%%%%%%%%%%%%%%%%%%%%%%%%%%%%%%%%%%%%%%%%%%%%%%%%%%%%%%%%%%%%%%%
%%%%%%%%%%%%%%%%%%%%%%%%%%%%%%%%%%%%%%%%%%%%%%%%%%%%%%%%%%%%%%%%%%
%\bigskip  % leave some space to appendix
\bigskip\bigskip  % leave more space
%\newpage  % new column actually
%\clearpage  % really a new page
\section*{Supplementary Material}
\appendix

\section{Additional Evaluation for the Direct Reflectance Prediction with a CNN (\cref{sec:direct_CNN_prediction})}

% Data Augmentation
%%%%%%%%%%
\subsection{Data Augmentation}\label{sec:augmentation}
As proposed in~\cite{zhou2015learningPriors}, we tried to augment the comparisons by computing the transitive closure of all comparisons.
Instead of pruning the comparisons with low confidence, as done in~\cite{zhou2015learningPriors}, we used all available annotations and set the weight $w_i$ for the augmented comparisons to be the minimum of the confidence of the pair of relations from which it was generated.
In case two relations for the same pair of points are generated, we keep the one with higher confidence.
In the end we do a consistency check and keep only consistent relations by throwing out the contradicting relation with lower confidence.
Despite the much bigger amount of data ($>20$M, a factor of 23.6 times as many comparisons), training on this augmented data did not improve on the resulting WHDR (computed on the original comparisons). %, but only slowed down the training time, since more comparisons have to be evaluated in each step.

%\tn{did you want to know the exact factor or the absolute number? we have 20,576,337 instead of 22,903,366 by Zhou (he does not throw out inconsistent ones)}

% Weak Label Analysis
%%%%%%%%%%
\subsection{Weak label analysis}\label{sec:weak_label}
\begin{figure}[t]
\centering
\includegraphics[width=0.9\linewidth]{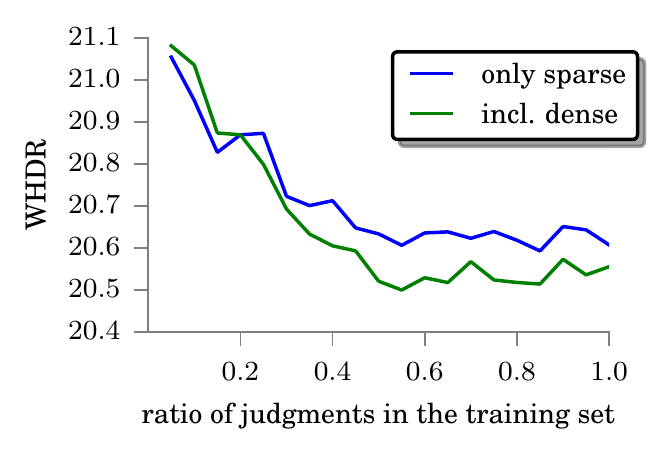}
\caption{
WHDR performance when training with fewer training annotations.
}
\label{fig:weak_label_analysis_line}
\end{figure}

We analyzed how much labeled information is needed to obtain good WHDR results.
To test this, we reduced the amount of available training data and retrained a fixed network with the parameters $n=5$, $f=2^5$, $k=1$, $\delta=0.12$, $\margin=0.08$ from scratch.
%\tn{This was with $n=3$, $f=2^5$, $k=3$. Currently re-running it with the chosen in the main submission. See also the comment in~\cref{fig:weak_label_analysis_line}}
First we reduce the amount of annotated pairs per image. The result is the green line in~\cref{fig:weak_label_analysis_line}.
We observe that it is possible to remove about $50\%$ of the annotation pairs until the WHDR loss starts to decrease.

Out of the 5230 images in IIW in total, roughly 400 images contain more ``dense'' annotations. This means they are evaluated at 303 to 1181 pairs (with a median of~916), instead of evaluating 1 to 216 (with a median of~108) comparisons.
When removing these images from the training set, the performance degrades~(see~blue  line in~\cref{fig:weak_label_analysis_line}), as expected.

\subsection{Network Hyper-Parameters}
A full parameter sweep of network hyper-parameters, over different kernel sizes, is given in~\cref{fig:numLayers_numFilters_kernels}. This is the result of training for more iterations and with a smaller batch size than in~\cref{fig:numLayers_numFilters}, to allow deeper nets with more convolutional filters. While performance does not differ much for $k=1$ and $k=3$, it seems that bigger kernel sizes overfit more heavily on the training set and therefore have higher mean WHDR on the validation set.

\def\size{0.65\columnwidth}
\begin{figure*}[t]
\centering
\subfigure[]{
\includegraphics[width=\size]{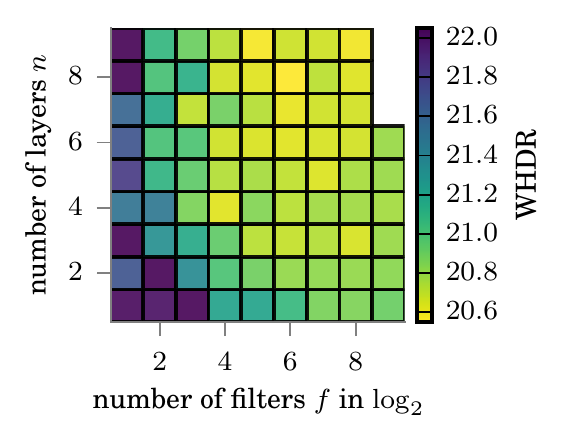}
\label{fig:numLayers_numFilters_k0}
}
\subfigure[]{
\includegraphics[width=\size]{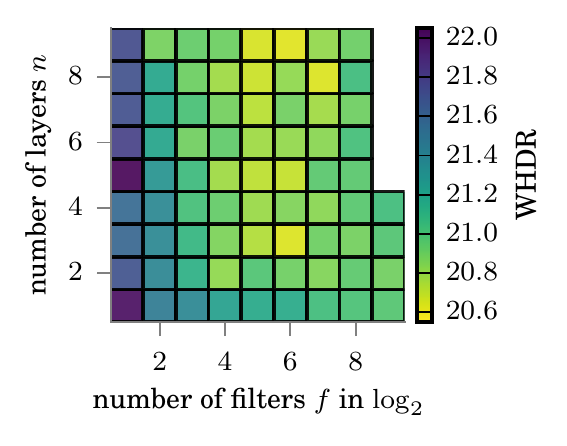}
\label{fig:numLayers_numFilters_k1}
}
\subfigure[]{
\includegraphics[width=\size]{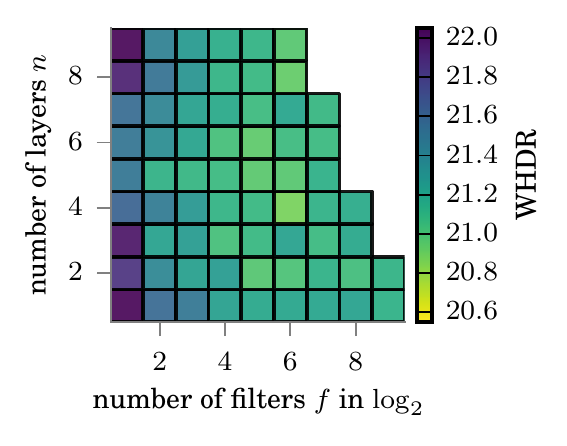}
\label{fig:numLayers_numFilters_k2}
}
\subfigure[]{
\includegraphics[width=\size]{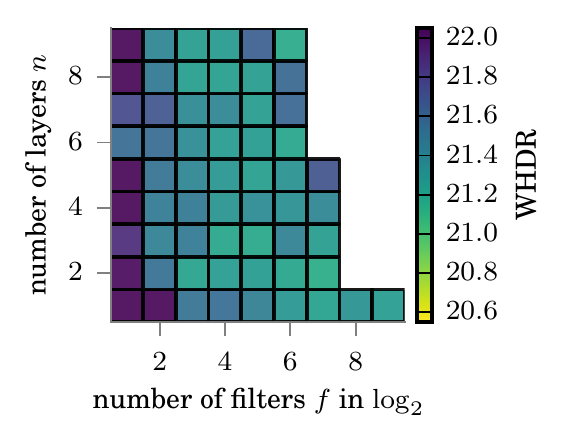}
\label{fig:numLayers_numFilters_k3}
}
\subfigure[]{
\includegraphics[width=\size]{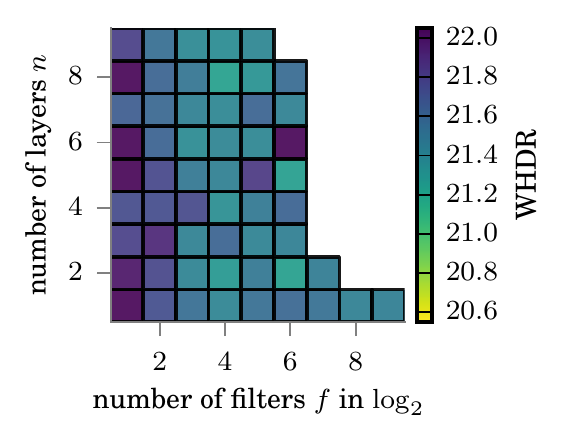}
\label{fig:numLayers_numFilters_k4}
}
\caption{Mean WHDR on the validation set for different network depths $n$ and number of filters $f$ for the kernel sizes
\subref{fig:numLayers_numFilters_k0}~$k=1$,
\subref{fig:numLayers_numFilters_k1}~$k=3$,
\subref{fig:numLayers_numFilters_k2}~$k=5$,
\subref{fig:numLayers_numFilters_k3}~$k=7$,
\subref{fig:numLayers_numFilters_k4}~$k=9$.
Missing data is the result of memory limit on our graphics card.
}
\label{fig:numLayers_numFilters_kernels}
\end{figure*}

%%%% lower bound scaling
\subsection{Rescaling of lower bound}
We give some more details what happens when a reflectance image is created by scaling the input image from~$[0,1]$ into~$[a,1]$, so that the lower bound has the constant value $a\in[0,1]$. Since WHDR measures reflectance ratios, the upper bound can be kept fixed to $1$ without loss of generality.
On the other hand, scaling the lower bound induces a non-linear change in the reflectance ratios, which influences the WHDR results.
For $a=0$ we have what~\cite{bell2014iiw} named baseline~(const~$S$), while $a=1$ ist baseline~(const~$R$).
Interestingly, using the parameter $a=0.55$, which gives the lowest WHDR on the training and validation set as shown in~\cref{fig:lower_bound_scaling:plot}, already outperforms Retinex, with a WHDR of 25.7 on the test set.
Our CNN that directly predicts dense reflectance, exploits this effect. This often leaves small variations in the reflectance image that should be explained via shading gradients, since they fall below the $\delta$ threshold for the ``equal'' class. It is this fact, which gave rise to smooth reflectance values by a filtering step to enforce piecewise constancy.

\bigskip

\section{Additional Evaluation for Reflectance Filtering (\cref{sec:filtering})}
\subsection{Hyper-Parameters of Filters}
For filtering the Direct CNN predictions with a bilateral filter, we searched for the spatial- and color scale hyper-parameters of the respective filter on the training and validation set in~\cref{fig:filtering:BF_ours_ours} to find $\sigma_s=22, \sigma_r=20$ having the lowest mean WHDR. On the test set, this improved the performance from $19.5$ to $18.9$.

Guided filtering also improved a bit to $19.2$ with $r=7$ and $\varepsilon=52$, chosen from~\cref{fig:filtering:GF_ours_ours}.

Dependence of WHDR performance on the validation set when filtering the direct CNN reflectance prediction with `flat' (see~\cref{sec:filtering_results}) is given in~\cref{fig:filtering:GF_ours_flat}. Taking $r=45$ and $\varepsilon=3$ leads to a test performance of $17.7$, on par with the current state-of-the art.

The filtering step is mostly dependent on the feature space, therefore we used the above filtering hyper-parameters when smoothing the method of Zoran et al. 2015~\cite{zoran2015ordinal}, since we only had access to their test set and hence could not optimize for the best parameters. Nonetheless using these parameters for filtering outperforms state-of-the-art with~$15.8$~mean WHDR by $1.9$ percentage points.

\def\size{0.61\columnwidth}
\begin{figure*}%[t]
\centering
%\subfigure[]{
%\includegraphics[width=\size]{filtering/joint_bilateral_filter_Input_with_Bi}
%\label{fig:filtering:BF_input_flat}
%}
\subfigure[]{
\includegraphics[width=0.41\columnwidth]{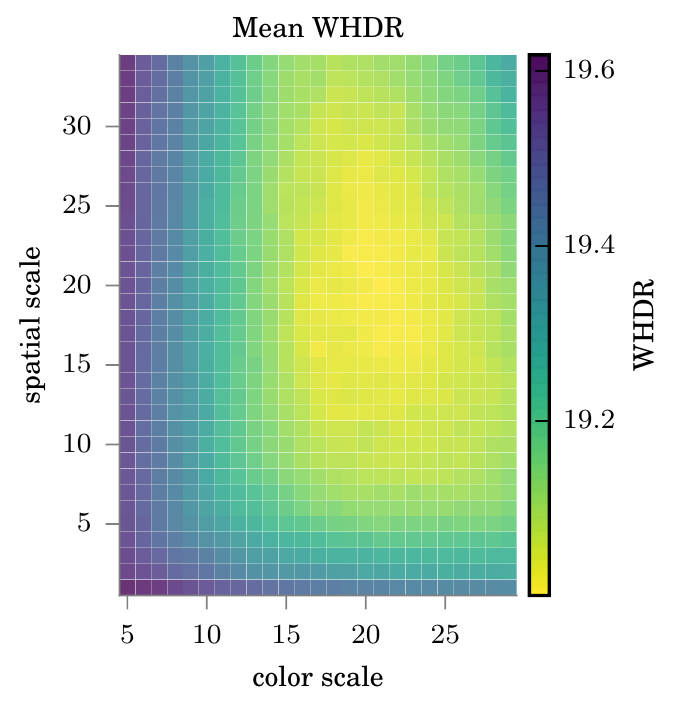}  % train + val
\label{fig:filtering:BF_ours_ours}
}
\subfigure[]{
\includegraphics[width=\size]{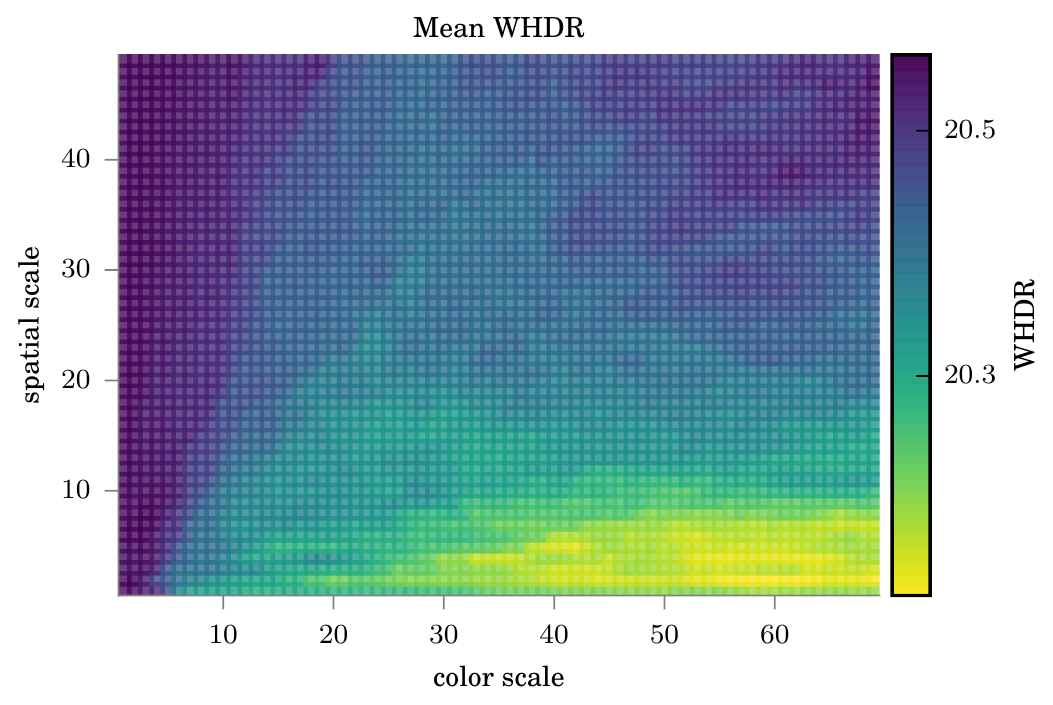}  % only val
\label{fig:filtering:GF_ours_ours}
}
\subfigure[]{
\includegraphics[width=\size]{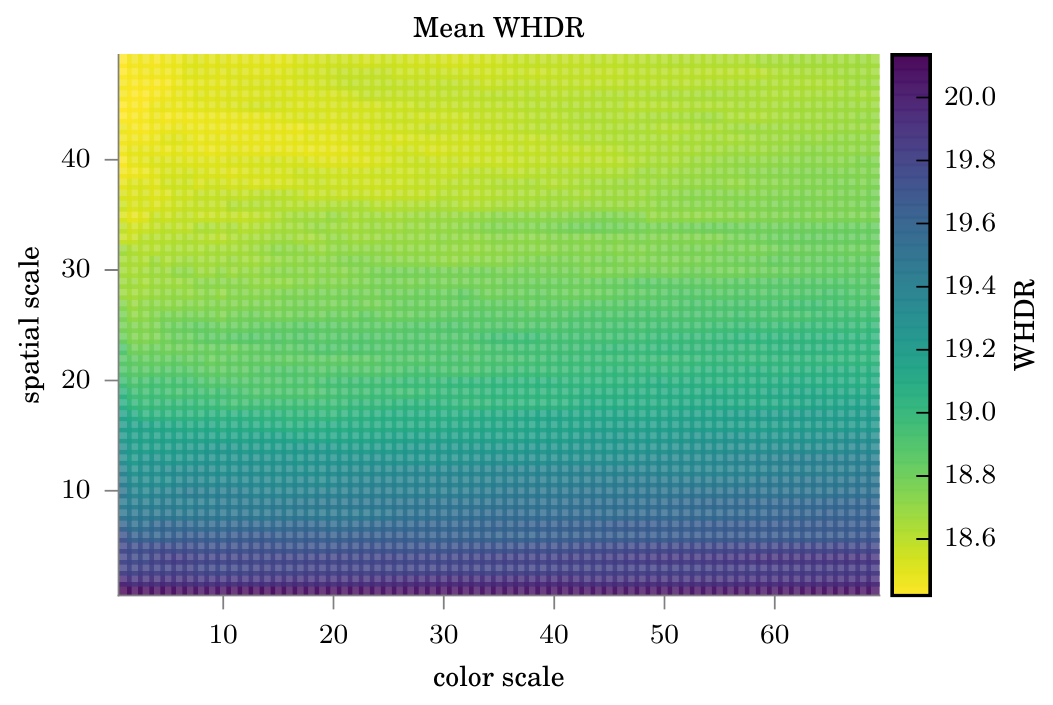}  % only val
\label{fig:filtering:GF_ours_flat}
}
\caption{Mean WHDR on the training and validation set for filtering
%\subref{fig:filtering:BF_input_flat}~BF(input, flat),
\subref{fig:filtering:BF_ours_ours}~BF(CNN, CNN),
\subref{fig:filtering:GF_ours_ours}~GF(CNN, CNN),
\subref{fig:filtering:GF_ours_flat}~GF(CNN, flat),
with a varying color and spatial scale.
}
\label{fig:filtering:sweeps}
\end{figure*}

\subsection{Extended Qualitative Results}
To assess the qualitative performance, we compiled a collection of results in~\cref{fig:qualitative_supplementary:first,fig:qualitative_supplementary:second,fig:qualitative_supplementary:third}.
The images are randomly sampled from the intersection of the Narihira~\cite{narihira2015lightness} and Zoran~\cite{zoran2015ordinal} test split.
The `flat' image used for guidance in filtering~(\cref{sec:filtering_results}) is given in the first row each.
In the spirit of the project page for~\cite{bell2014iiw} we also show grayscale reflectance, especially to highlight the difference between the baseline (const $R$) and our direct CNN reflectance prediction, which appears to be subtle in the color reflectance, but is not to be overlooked in the grayscale reflectance. The method of Zoran et al.~\cite{zoran2015ordinal} has staircase effects due to the superpixelization in reflectance. This is removed by our reflectance filtering step, when filtering with the flat image, which not only leads to improved quantitative, but also qualitative results.

%%%%% qualitative results

%\def\ourMethod{nestmeyer2016}
\def\ourMethod{nestmeyer2016_whdrCNN_1004_skipLayers_n3_f32_delta_margin_14_08_then_13_09}

% other available IDs already in the figures folder: '611', '100585', '100091', '91559', '62440', '116625'

% for new submission
\def\photoIDa{101684}
\def\photoIDb{102147}

\def\methoda{baseline_reflectance}
\def\methodaName{Baseline (const $R$)}

\def\methodb{nestmeyer2016_whdrCNN_1109_rDirectly_from_1108_single_channel_wdm_12_06_then_13_08_linear_colorized_sRGB}
\def\methodbName{\textbf{CNN prediction}}

\def\methodc{ours_guided_c3.0s45.0_bi_flat_linear_colorized_sRGB}
\def\methodcName{\textbf{GF(CNN, flat)}}

\def\methodd{bi2015_l1_linear_colorized_sRGB}
\def\methoddName{Bi et al. 2015~\cite{bi2015l1}}

\def\methode{bi2015_l1_final_linear_smoothed_with_Bi_flat_linear_colorized_sRGB}
\def\methodeName{\textbf{BF(\cite{bi2015l1}, flat)}}

\def\methodf{zoran2015_ordinal_onlyHisTest_linear_colorized_sRGB}
\def\methodfName{Zoran et al. 2015~\cite{zoran2015ordinal}}

\def\methodg{zoran_guided_c3.0s45.0_bi_flat_linear_guided_c3.0s45.0_bi_flat_linear_guided_c3.0s45.0_bi_flat_linear_colorized_sRGB}
\def\methodgName{\textbf{3xGF(\cite{zoran2015ordinal}, flat)}}
\def\size{0.128\textwidth}  % width of all images shown below
\def\nameSize{16mm}  % for 6 images per row
\begin{figure*}
%\vspace{-5mm}
\centering
\setlength\extrarowheight{-3pt}  % less space is used up to -2, then no further improvement
\begin{tabular}{cccc|ccc}
 & \multicolumn{3}{c}{Photo ID \photoIDa} & \multicolumn{3}{c}{Photo ID \photoIDb} \\
 & input image & human judgments & flat &
   input image & human judgments & flat\\
%\rotatebox[origin=lc]{90}{\begin{minipage}{\nameSize}\raggedright{input and judgments}\end{minipage}}
 &
\includegraphics[width=\size]{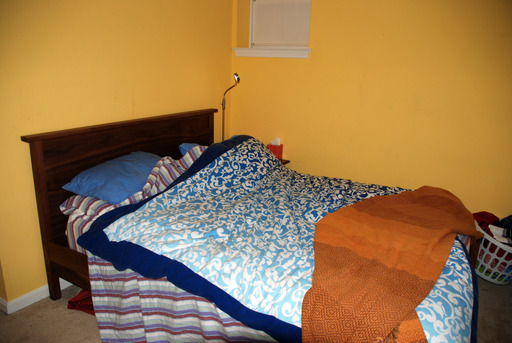} &
\includegraphics[width=\size]{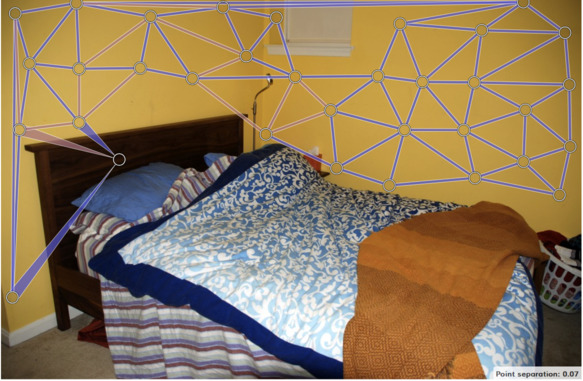} &
\includegraphics[width=\size]{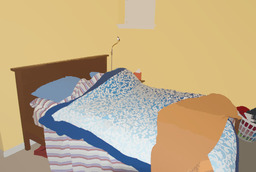} &
\includegraphics[width=\size]{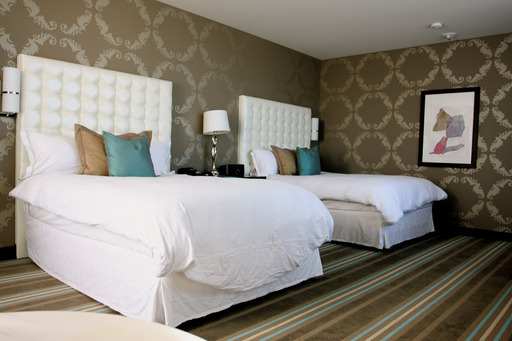} &
\includegraphics[width=\size]{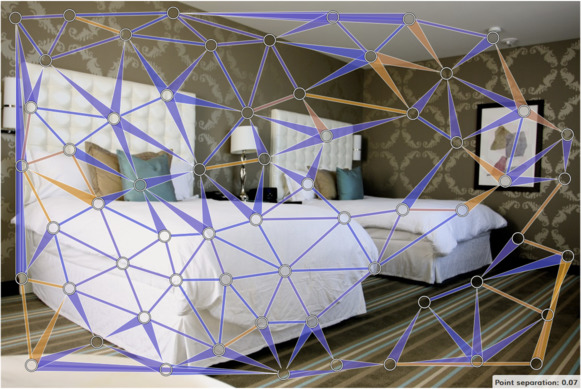} &
\includegraphics[width=\size]{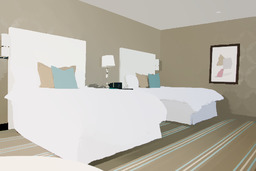} \\
 &
reflectance & grayscale refl. & shading &
reflectance & grayscale refl. & shading \\
% ours
\rotatebox[origin=lc]{90}{\begin{minipage}{\nameSize}\raggedright{\methodaName}\end{minipage}} &
\includegraphics[width=\size]{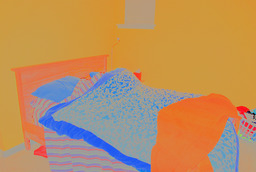} &
\includegraphics[width=\size]{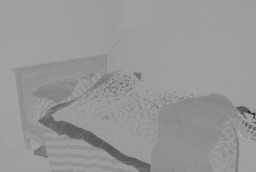} &
\includegraphics[width=\size]{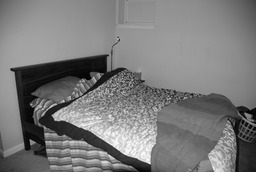} &
\includegraphics[width=\size]{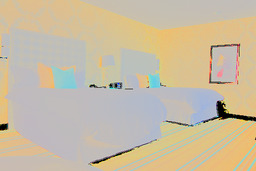} &
\includegraphics[width=\size]{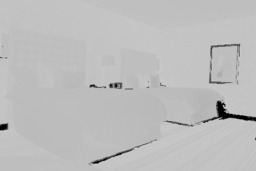} &
\includegraphics[width=\size]{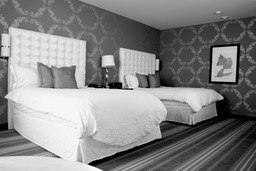} \\
\rotatebox[origin=lc]{90}{\begin{minipage}{\nameSize}\raggedright{\methodbName}\end{minipage}} &
\includegraphics[width=\size]{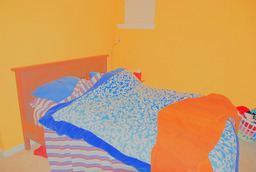} &
\includegraphics[width=\size]{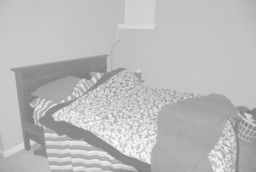} &
\includegraphics[width=\size]{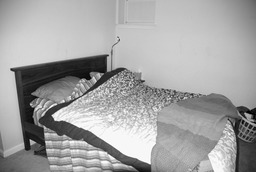} &
\includegraphics[width=\size]{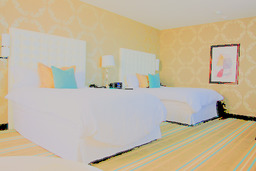} &
\includegraphics[width=\size]{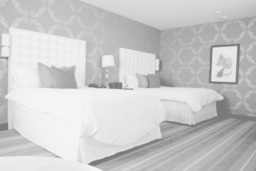} &
\includegraphics[width=\size]{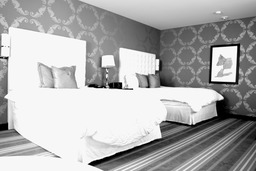} \\
\rotatebox[origin=lc]{90}{\begin{minipage}{\nameSize}\raggedright{\methodcName}\end{minipage}} &
\includegraphics[width=\size]{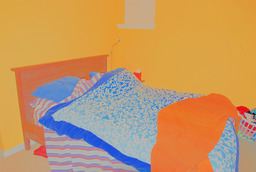} &
\includegraphics[width=\size]{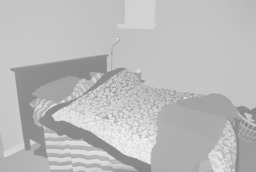} &
\includegraphics[width=\size]{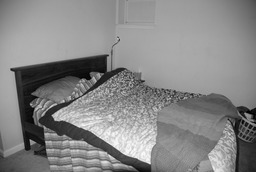} &
\includegraphics[width=\size]{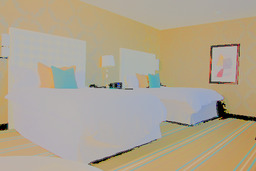} &
\includegraphics[width=\size]{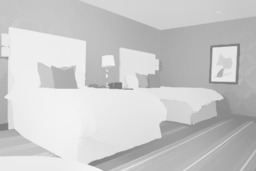} &
\includegraphics[width=\size]{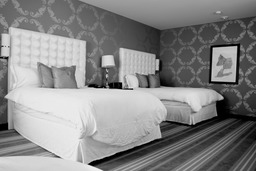} \\
\rotatebox[origin=lc]{90}{\begin{minipage}{\nameSize}\raggedright{\methoddName}\end{minipage}} &
\includegraphics[width=\size]{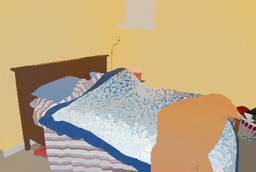} &
\includegraphics[width=\size]{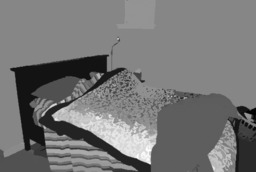} &
\includegraphics[width=\size]{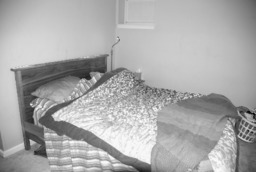} &
\includegraphics[width=\size]{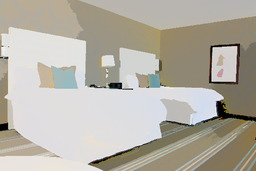} &
\includegraphics[width=\size]{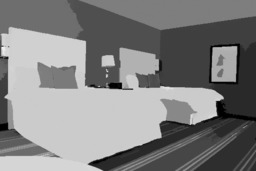} &
\includegraphics[width=\size]{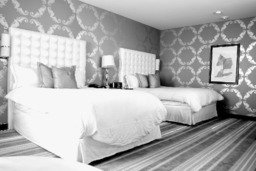} \\
\rotatebox[origin=lc]{90}{\begin{minipage}{\nameSize}\raggedright{\methodeName}\end{minipage}} &
\includegraphics[width=\size]{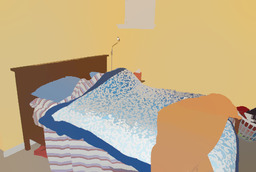} &
\includegraphics[width=\size]{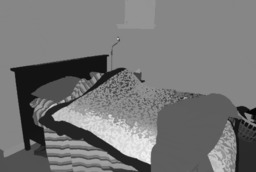} &
\includegraphics[width=\size]{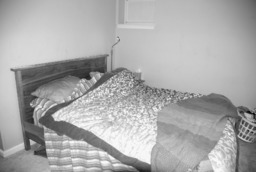} &
\includegraphics[width=\size]{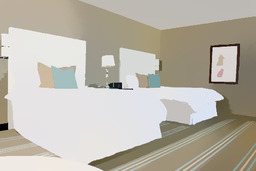} &
\includegraphics[width=\size]{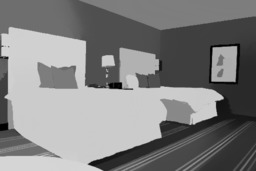} &
\includegraphics[width=\size]{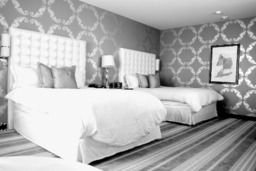} \\
\rotatebox[origin=lc]{90}{\begin{minipage}{\nameSize}\raggedright{\methodfName}\end{minipage}} &
\includegraphics[width=\size]{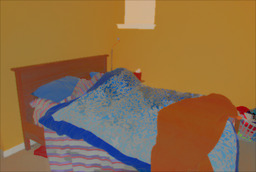} &
\includegraphics[width=\size]{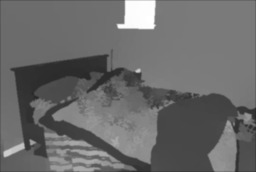} &
\includegraphics[width=\size]{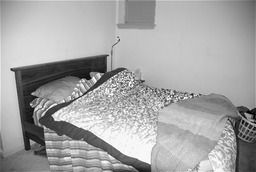} &
\includegraphics[width=\size]{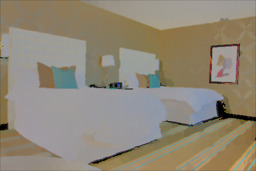} &
\includegraphics[width=\size]{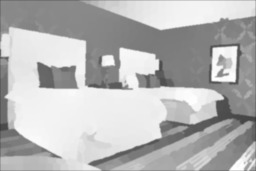} &
\includegraphics[width=\size]{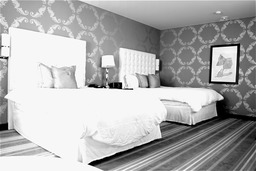} \\
\rotatebox[origin=lc]{90}{\begin{minipage}{\nameSize}\raggedright{\methodgName}\end{minipage}} &
\includegraphics[width=\size]{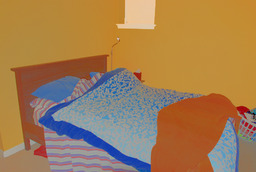} &
\includegraphics[width=\size]{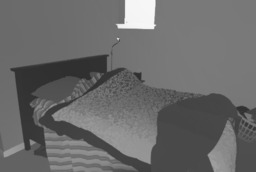} &
\includegraphics[width=\size]{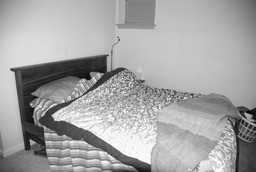} &
\includegraphics[width=\size]{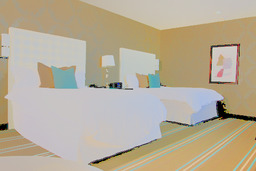} &
\includegraphics[width=\size]{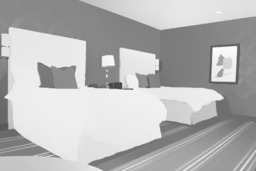} &
\includegraphics[width=\size]{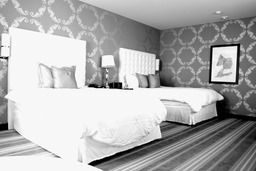}
\end{tabular}
\caption{Qualitative comparison on sample images of IIW.
The first row gives the input image, the evaluated comparisons on it and the flat image (see~\cref{sec:filtering_results}) used for filtering.
Comparisons are given as in~\cite{bell2014iiw}, where blue is a judgment with high confidence and orange low. The narrow part of the connecting lines is the point which is labeled as darker or they are given as ``about the same'' when the annotation is a straight line.
In the following rows the decompositions into color reflectance in the first column, grayscale reflectance in the second and shading in the third of a subset of methods is shown. All outputs are mapped to sRGB for display.
%\tn{It might make sense to report WHDR for each method on the given image? But that will take a while to produce.}
}
\label{fig:qualitative_supplementary:first}
\end{figure*}

%%%%%%%%%%%%%%%%%%%%%%%%%%%%%%%%%%%%%%%%%%%%%%%%%%%%%%%%%%%%%%%%%%%%%%%%%%%%%%%%%%%%%

\def\photoIDa{78671}
\def\photoIDb{9499}

\begin{figure*}
%\vspace{-5mm}
\centering
\setlength\extrarowheight{-3pt}  % less space is used up to -2, then no further improvement
\begin{tabular}{cccc|ccc}
 & \multicolumn{3}{c}{Photo ID \photoIDa} & \multicolumn{3}{c}{Photo ID \photoIDb} \\
 & input image & human judgments & flat &
   input image & human judgments & flat\\
%\rotatebox[origin=lc]{90}{\begin{minipage}{\nameSize}\raggedright{input and judgments}\end{minipage}}
 &
\includegraphics[width=\size]{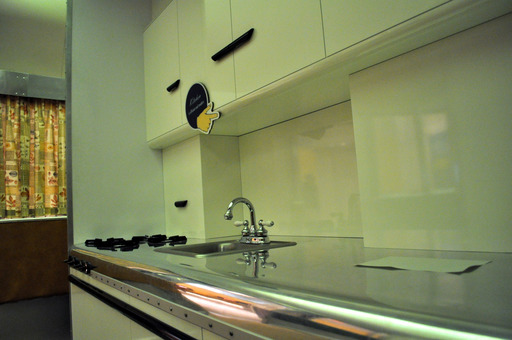} &
\includegraphics[width=\size]{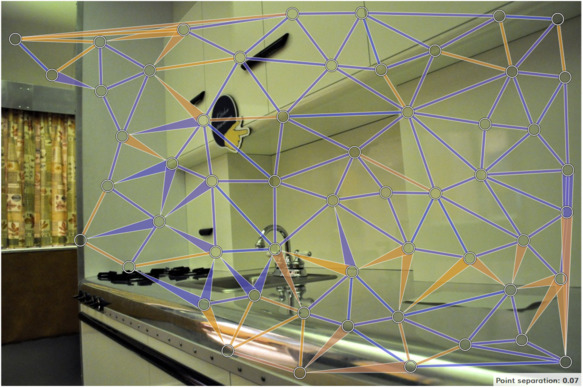} &
\includegraphics[width=\size]{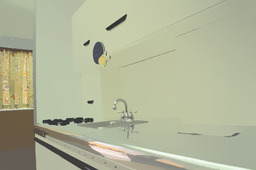} &
\includegraphics[width=\size]{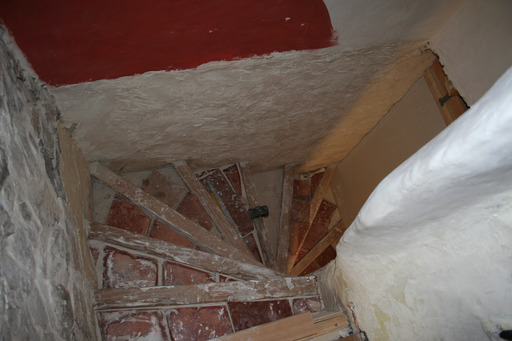} &
\includegraphics[width=\size]{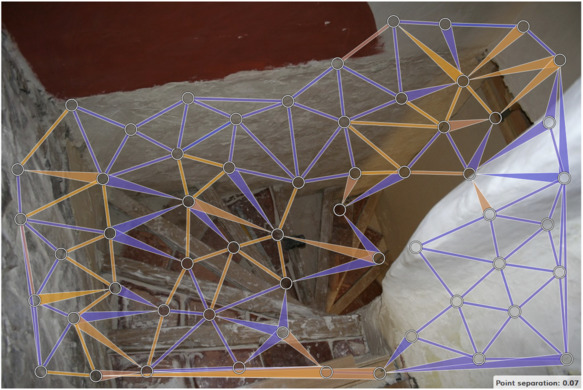} &
\includegraphics[width=\size]{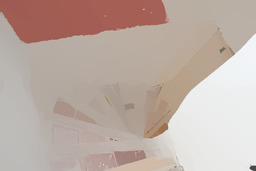} \\
 &
reflectance & grayscale refl. & shading &
reflectance & grayscale refl. & shading \\
% ours
\rotatebox[origin=lc]{90}{\begin{minipage}{\nameSize}\raggedright{\methodaName}\end{minipage}} &
\includegraphics[width=\size]{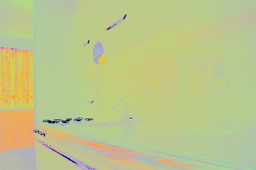} &
\includegraphics[width=\size]{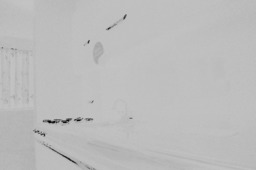} &
\includegraphics[width=\size]{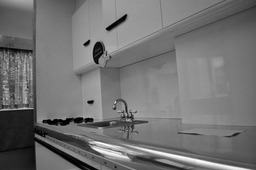} &
\includegraphics[width=\size]{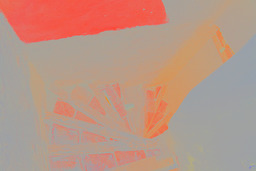} &
\includegraphics[width=\size]{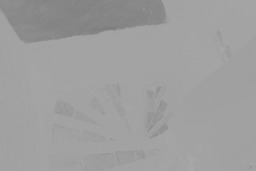} &
\includegraphics[width=\size]{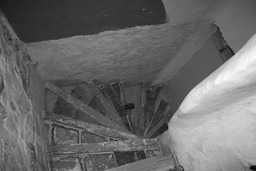} \\
\rotatebox[origin=lc]{90}{\begin{minipage}{\nameSize}\raggedright{\methodbName}\end{minipage}} &
\includegraphics[width=\size]{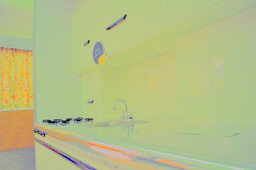} &
\includegraphics[width=\size]{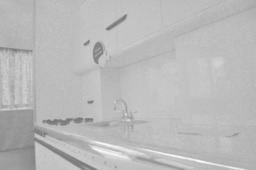} &
\includegraphics[width=\size]{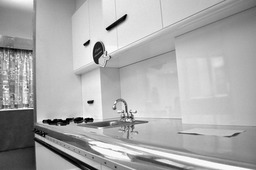} &
\includegraphics[width=\size]{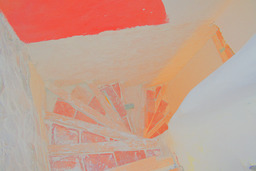} &
\includegraphics[width=\size]{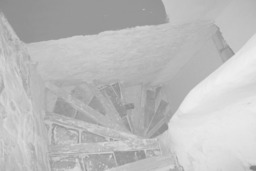} &
\includegraphics[width=\size]{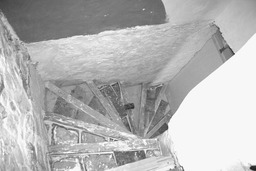} \\
\rotatebox[origin=lc]{90}{\begin{minipage}{\nameSize}\raggedright{\methodcName}\end{minipage}} &
\includegraphics[width=\size]{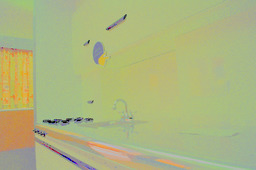} &
\includegraphics[width=\size]{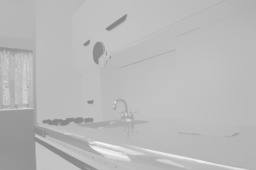} &
\includegraphics[width=\size]{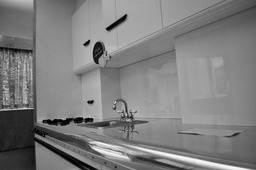} &
\includegraphics[width=\size]{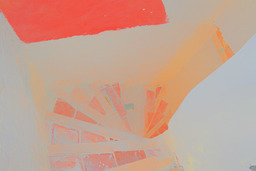} &
\includegraphics[width=\size]{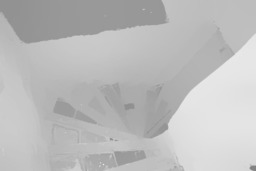} &
\includegraphics[width=\size]{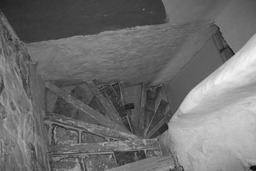} \\
\rotatebox[origin=lc]{90}{\begin{minipage}{\nameSize}\raggedright{\methoddName}\end{minipage}} &
\includegraphics[width=\size]{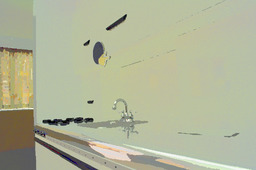} &
\includegraphics[width=\size]{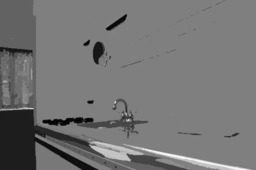} &
\includegraphics[width=\size]{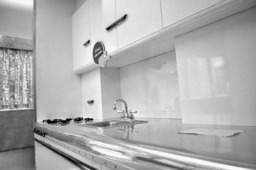} &
\includegraphics[width=\size]{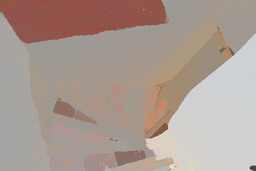} &
\includegraphics[width=\size]{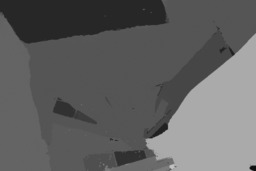} &
\includegraphics[width=\size]{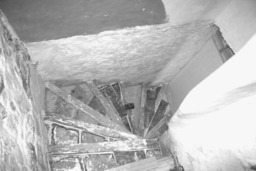} \\
\rotatebox[origin=lc]{90}{\begin{minipage}{\nameSize}\raggedright{\methodeName}\end{minipage}} &
\includegraphics[width=\size]{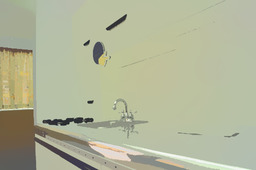} &
\includegraphics[width=\size]{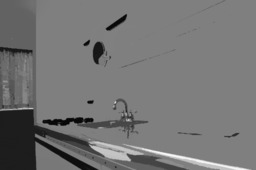} &
\includegraphics[width=\size]{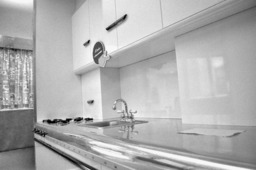} &
\includegraphics[width=\size]{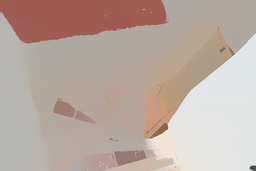} &
\includegraphics[width=\size]{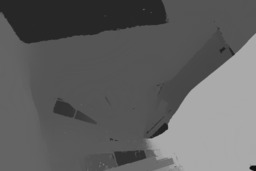} &
\includegraphics[width=\size]{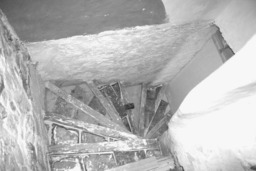} \\
\rotatebox[origin=lc]{90}{\begin{minipage}{\nameSize}\raggedright{\methodfName}\end{minipage}} &
\includegraphics[width=\size]{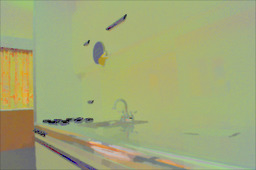} &
\includegraphics[width=\size]{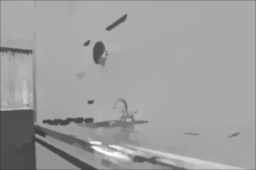} &
\includegraphics[width=\size]{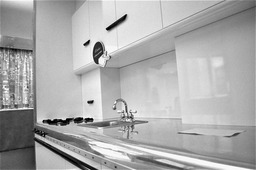} &
\includegraphics[width=\size]{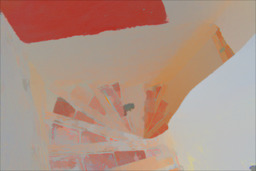} &
\includegraphics[width=\size]{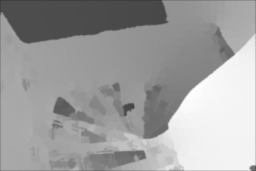} &
\includegraphics[width=\size]{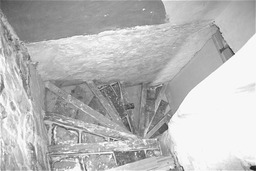} \\
\rotatebox[origin=lc]{90}{\begin{minipage}{\nameSize}\raggedright{\methodgName}\end{minipage}} &
\includegraphics[width=\size]{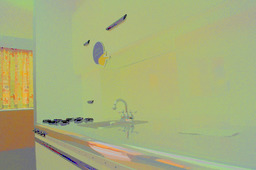} &
\includegraphics[width=\size]{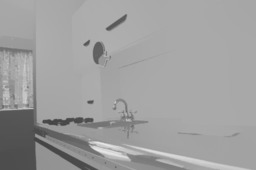} &
\includegraphics[width=\size]{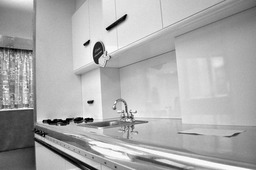} &
\includegraphics[width=\size]{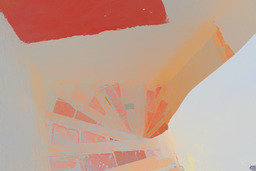} &
\includegraphics[width=\size]{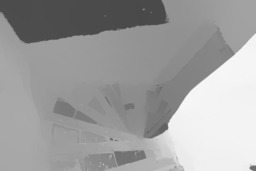} &
\includegraphics[width=\size]{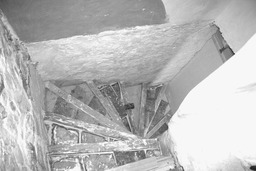}
\end{tabular}
\caption{Extends~\cref{fig:qualitative_supplementary:first} with Photo IDs \photoIDa{} and \photoIDb{}.}
\label{fig:qualitative_supplementary:second}
\end{figure*}

%%%%%%%%%%%%%%%%%%%%%%%%%%%%%%%%%%%%%%%%%%%%%%%%%%%%%%%%%%%%%%%%%%%%%%%%%%%%%%%%%%%%%

\def\photoIDa{60820}
\def\photoIDb{34647}

\begin{figure*}
%\vspace{-5mm}
\centering
\setlength\extrarowheight{-3pt}  % less space is used up to -2, then no further improvement
\begin{tabular}{cccc|ccc}
 & \multicolumn{3}{c}{Photo ID \photoIDa} & \multicolumn{3}{c}{Photo ID \photoIDb} \\
 & input image & human judgments & flat &
   input image & human judgments & flat\\
%\rotatebox[origin=lc]{90}{\begin{minipage}{\nameSize}\raggedright{input and judgments}\end{minipage}}
 &
\includegraphics[width=\size]{decompositions/input/\photoIDa-input} &
\includegraphics[width=\size]{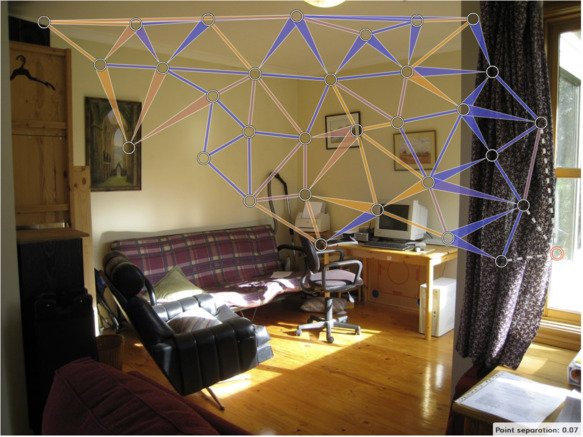} &
\includegraphics[width=\size]{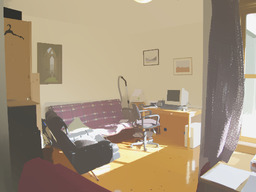} &
\includegraphics[width=\size]{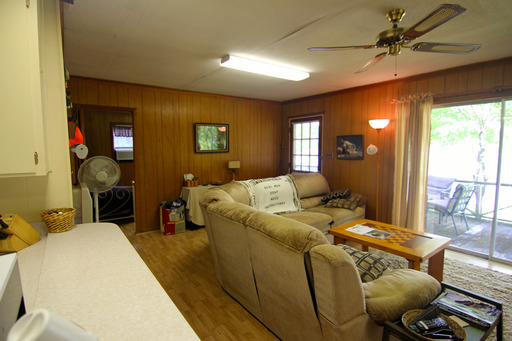} &
\includegraphics[width=\size]{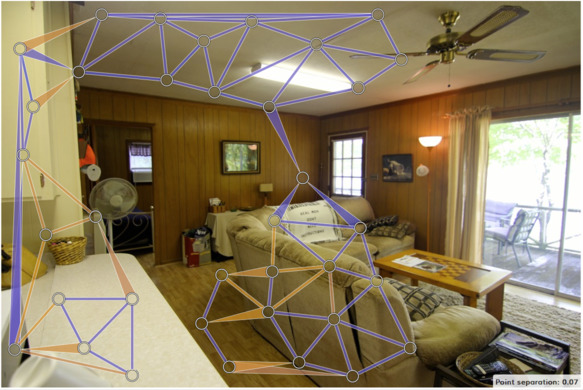} &
\includegraphics[width=\size]{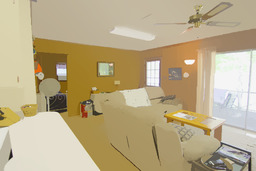} \\
 &
reflectance & grayscale refl. & shading &
reflectance & grayscale refl. & shading \\
% ours
\rotatebox[origin=lc]{90}{\begin{minipage}{\nameSize}\raggedright{\methodaName}\end{minipage}} &
\includegraphics[width=\size]{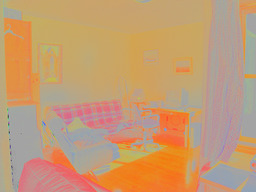} &
\includegraphics[width=\size]{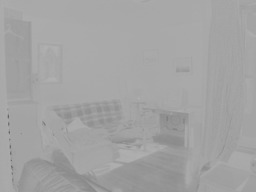} &
\includegraphics[width=\size]{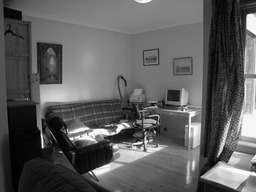} &
\includegraphics[width=\size]{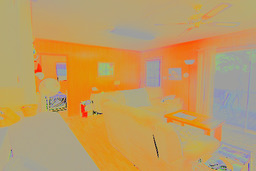} &
\includegraphics[width=\size]{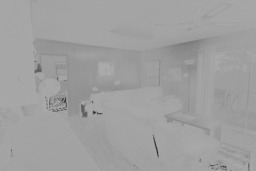} &
\includegraphics[width=\size]{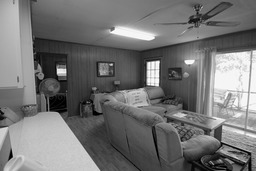} \\
\rotatebox[origin=lc]{90}{\begin{minipage}{\nameSize}\raggedright{\methodbName}\end{minipage}} &
\includegraphics[width=\size]{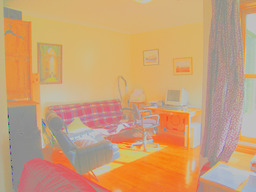} &
\includegraphics[width=\size]{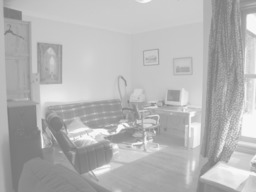} &
\includegraphics[width=\size]{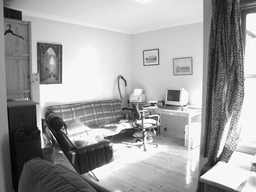} &
\includegraphics[width=\size]{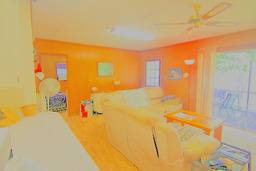} &
\includegraphics[width=\size]{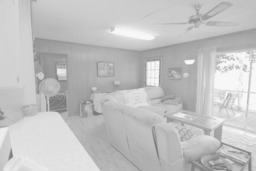} &
\includegraphics[width=\size]{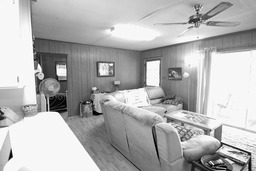} \\
\rotatebox[origin=lc]{90}{\begin{minipage}{\nameSize}\raggedright{\methodcName}\end{minipage}} &
\includegraphics[width=\size]{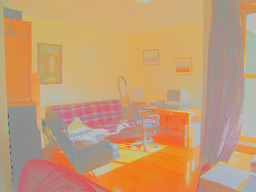} &
\includegraphics[width=\size]{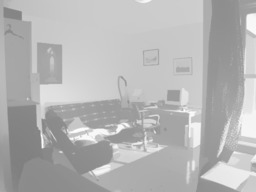} &
\includegraphics[width=\size]{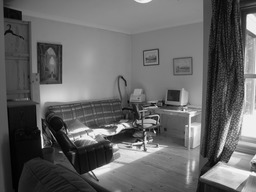} &
\includegraphics[width=\size]{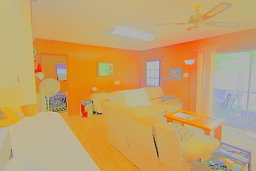} &
\includegraphics[width=\size]{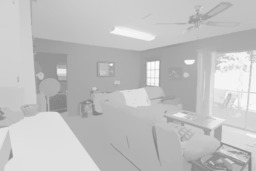} &
\includegraphics[width=\size]{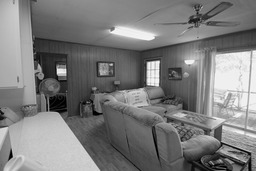} \\
\rotatebox[origin=lc]{90}{\begin{minipage}{\nameSize}\raggedright{\methoddName}\end{minipage}} &
\includegraphics[width=\size]{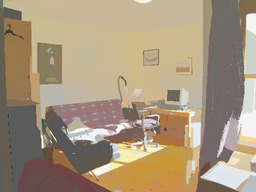} &
\includegraphics[width=\size]{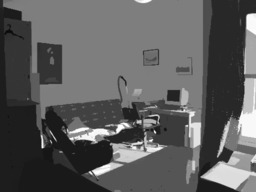} &
\includegraphics[width=\size]{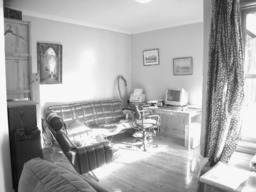} &
\includegraphics[width=\size]{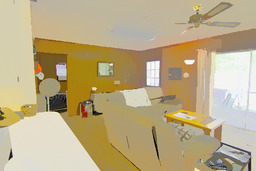} &
\includegraphics[width=\size]{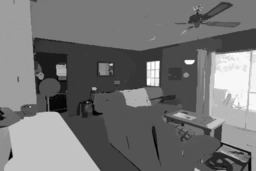} &
\includegraphics[width=\size]{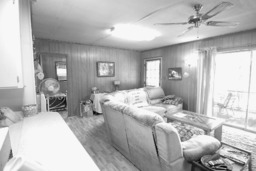} \\
\rotatebox[origin=lc]{90}{\begin{minipage}{\nameSize}\raggedright{\methodeName}\end{minipage}} &
\includegraphics[width=\size]{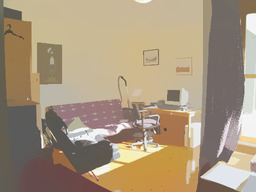} &
\includegraphics[width=\size]{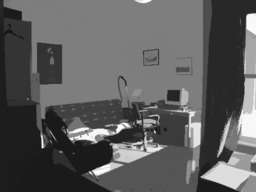} &
\includegraphics[width=\size]{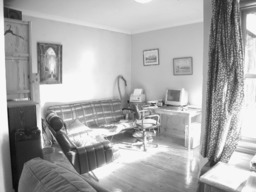} &
\includegraphics[width=\size]{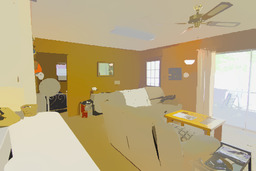} &
\includegraphics[width=\size]{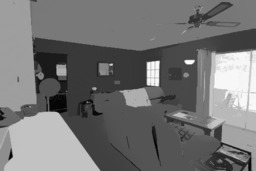} &
\includegraphics[width=\size]{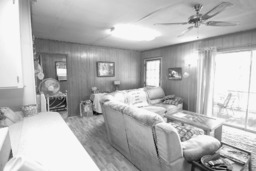} \\
\rotatebox[origin=lc]{90}{\begin{minipage}{\nameSize}\raggedright{\methodfName}\end{minipage}} &
\includegraphics[width=\size]{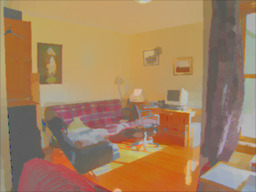} &
\includegraphics[width=\size]{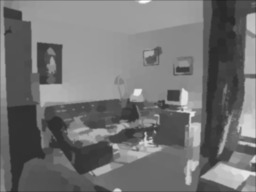} &
\includegraphics[width=\size]{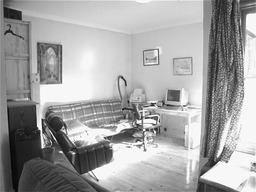} &
\includegraphics[width=\size]{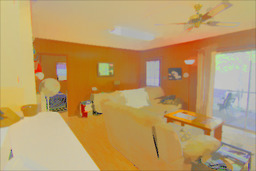} &
\includegraphics[width=\size]{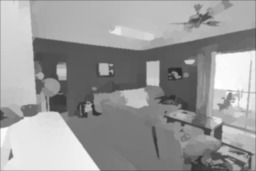} &
\includegraphics[width=\size]{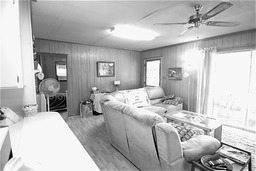} \\
\rotatebox[origin=lc]{90}{\begin{minipage}{\nameSize}\raggedright{\methodgName}\end{minipage}} &
\includegraphics[width=\size]{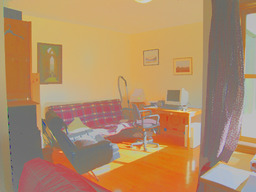} &
\includegraphics[width=\size]{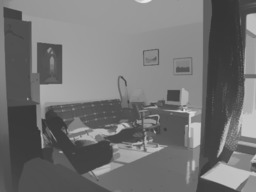} &
\includegraphics[width=\size]{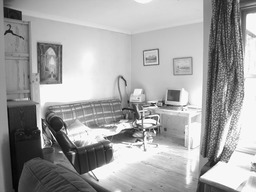} &
\includegraphics[width=\size]{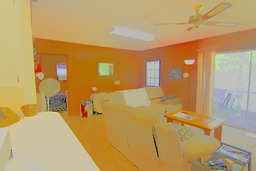} &
\includegraphics[width=\size]{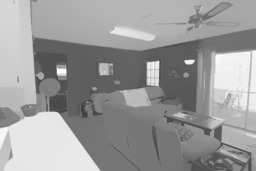} &
\includegraphics[width=\size]{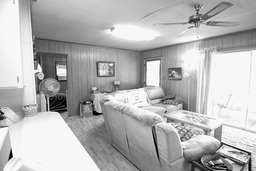}
\end{tabular}
\caption{Extends~\cref{fig:qualitative_supplementary:first} with Photo IDs \photoIDa{} and \photoIDb{}.}
\label{fig:qualitative_supplementary:third}
\end{figure*}

\end{document}